\documentclass{article}

\usepackage[final]{neurips_data_2024}

\usepackage[utf8]{inputenc} % allow utf-8 input
\usepackage[T1]{fontenc}    % use 8-bit T1 fonts
\usepackage{hyperref}       % hyperlinks
\usepackage{url}            % simple URL 
\usepackage{booktabs}       % professional-quality tables
\usepackage{amsfonts}       % blackboard math symbols
\usepackage{nicefrac}       % compact symbols for 1/2, etc.
\usepackage{microtype}      % microtypography
\usepackage{xcolor}         % colors
\usepackage{multirow}
\usepackage{diagbox}
\usepackage{listings}
\usepackage{comment}

\usepackage{utfsym}
\newcommand{\crossmark}{\scalebox{0.75}{\usym{2613}}}

\usepackage{float}
\usepackage{tabularx}

\usepackage{siunitx}

\usepackage{graphicx}

\usepackage{cleveref}

\usepackage[nolist]{acronym}
\begin{acronym}
\acro{CNN}{Convolutional Neural Network}
\acro{DL}{Deep Learning}
\acro{DSC}{Dice similarity coefficient}
\acro{GAN}{Generative Adversarial Network}
\acro{ML}{Machine Learning}
\acro{GT}{Ground Truth}
\acro{PGT}{Peak Ground Truth}
\acro{RWMP}{Real World Model Performance}
\acro{NN}{Neural Network}
\acro{SOTA}{State-of-the-Art}
\acro{RPN}{Region Proposal Network}
\acro{P-R}{Precision-Recall}
\acro{TP}{True Positive}
\acro{FP}{False Positive}
\acro{FN}{False Negative}
\acro{TPs}{True Positives}
\acro{FPs}{False Positives}
\acro{FNs}{False Negatives}
\acro{mAP}{Mean Average Precision}
\acro{AP}{Average Precision}
\acro{IOU}{Intersection-over-Union}
\acro{BST}{Box score threshold}
\acro{NMS}{Non max suppression}
\acro{SSD}{Single Shot MultiBox Detector} 
\acro{YOLOv3}{You Only Look Once, Version 3} 
\acro{RTMDet}{Real-Time Models for object Detection} 
\end{acronym}

\newcommand{\eac}[1]{\emph{\ac{#1}}}

\usepackage{placeins}

\title{MultiOrg: A Multi-rater Organoid-detection Dataset}

\author{%
Christina Bukas$^1$, Harshavardhan Subramanian$^1$, Fenja See$^2$, Carina Steinchen$^2$\\
\textbf{Ivan Ezhov}$^3$, \textbf{Gowtham Boosarpu}$^{4}$, \textbf{Sara Asgharpour}$^2$, \textbf{Gerald Burgstaller}$^2$\\
\textbf{Mareike Lehmann}$^{2,4}$, \textbf{Florian Kofler}$^{1,5,6}$, \textbf{Marie Piraud}$^1$\\
$^1$Helmholtz AI, Computational Health Center (CHC), Helmholtz Munich, Neuherberg, Germany\\
$^2$Institute of Lung Health and Immunity (LHI), Comprehensive Pneumology Center (CPC),\\
Helmholtz Munich, Member of the German Center for Lung Research (DZL), Neuherberg, Germany\\
$^3$Technical University of Munich, School of Computation, Information and Technology,\\
Department of Computer Science, Munich, Germany\\
$^4$Institute for Lung Research, Philipps-University Marburg,
Universities of Giessen and\\
Marburg Lung Center, Member of the German Center for Lung Research (DZL), Marburg, Germany\\
$^5$ Department of Neuroradiology, Technical University of Munich, Munich, Germany. \\
$^6$ Department of Quantitative Biomedicine, University of
Zurich, Switzerland. \\
\texttt{\{christina.bukas,mareike.lehmann,marie.piraud\}@helmholtz-munich.de}\\
}

\begin{document}

\maketitle

\begin{abstract}

High-throughput image analysis in the biomedical domain has gained significant attention in recent years, driving advancements in drug discovery, disease prediction, and personalized medicine.
Organoids, specifically, are an active area of research, providing excellent models for human organs and their functions.
Automating the quantification of organoids in microscopy images would provide an effective solution to overcome substantial manual quantification bottlenecks, particularly in high-throughput image analysis.
However, there is a notable lack of open biomedical datasets, % dedicated to organoid detection and tracking, 
in contrast to other domains, such as autonomous driving, and, notably, only few of them have attempted to quantify annotation uncertainty.
In this work, we present \emph{MultiOrg} a comprehensive organoid dataset tailored for object detection tasks with uncertainty quantification.
This dataset comprises over 400 high-resolution 2d microscopy images and curated annotations of more than 60,000 organoids.
Most importantly, it includes three label sets for the test data, independently annotated by two experts at distinct time points.
We additionally provide a benchmark for organoid detection, and make the best model available through an easily installable, interactive plugin for the popular image visualization tool Napari, % which leverages a deep learning architecture 
to perform organoid quantification.
% and report the performance of four models on the available annotations. %and report a performance of 0.738 recall and 0.44 precision on the test set for a standard box score threshold of 0.5.
% We also present an easily installable, interactive plugin for the popular image visualization tool napari, which leverages a deep learning architecture to perform the quantification task.
% To the best of our knowledge, this represents the most extensive organoid dataset for a specific organ—lungs. Moreover, our primary objective in releasing this meticulously curated dataset to the scientific community is to expedite organoid research and enable its utilization in various object detection and downstream applications within the biomedical domain.
\end{abstract}

% The abstract paragraph should be indented 1/2 inch (3 picas) on both the left- and right-hand margins. Use 10 point type, with a vertical spacing (leading) of 11 points. The word Abstract must be centered, bold, and in point size 12. Two line spaces precede the abstract. The abstract must be limited to one paragraph.

% \input{flow/latex_evangelium}
\section{Introduction}
\label{sec:intro}

% WHY ORGANOIDS MATTER
% In biomedical image analysis, the imperative for accurate and efficient object detection methodologies underscores a critical challenge with far-reaching implications for diagnostics and research. This requires a meticulously curated dataset encompassing a diverse array of high-resolution medical images, mirroring the complexities of real-world scenarios (ex., label-noise awareness). 
Accurate and efficient object detection methods in biomedical image analysis are crucial for research and diagnostics.
% This requires diverse, well-curated datasets of high-resolution images reflecting real-world complexities, \textcolor{red}{which are rare in  this field}.
% This work presents a multi-rater organoid dataset designed for benchmarking object detection algorithms.
% 
Designing such methods requires diverse, well-curated datasets of high-resolution images reflecting real-world complexities.
The annotation of biomedical datasets represents a labor-intensive and subjective process relying on human experts.
This work represents a multi-rater organoid dataset designed for benchmarking object detection algorithms in a label-noise-aware setting that embraces the subjectivity in labels.

Organoids are miniature three-dimensional (3d) models of organs grown in vitro from stem cells.
They mimic the complexity and functionality of real organs, making them extremely valuable for medical research, disease modeling, and drug testing \citep{barkauskas2017lung, kim2020human, ingber2022human}.
%
% Organoids are self-assembling three-dimensional spheres, growing in a gel-like environment. Depending on the seeded progenitor cells from which they derive, organoids can mimic different organs and organ systems. Compared to using simple cell culture in biological experiments, the organoid model has the unique advantage of studying the organic development, interaction between different cell types, and behavior of stem cells ex-vivo in a 3D environment.
Organoid cultures, deriving from healthy and diseased or genetically engineered cells and undergoing different conditions and treatments, can be grown for several months \citep{youk2020three, huch2015modeling}.
These high-throughput experiments are monitored via microscopic imaging and, therefore, necessitate fast and objective detection, quantification, and tracking methods \citep{rios2018imaging, du2023organoids}.
% Furthermore, organoids can be imaged, stained for specific expression markers, passaged, and cultured for several months \citep{youk2020three} \citep{huch2015modeling}. The detection and tracking of organoids allows us to distinguish the effects of different treatments or cell types derived from the seeded progenitor cell culture. One can conclude regenerative capacity and plasticity by evaluating the colony-forming efficiency and sizes of organoids. Moreover, deferentially sized organoids can represent different organoid types. Therefore, it is of great importance to have an objective quantification method of organoid counts to prevent as many biases as humanly possible.
%
Detection of organoids in real-world lab-culture images is associated with many challenges \citep{kassis2019orgaquant}.
Beyond the typical challenges associated with microscopy (out-of-focus, lightning, padding, etc...), those 3d cultures are imaged in 2d, leading to overlapping structures.
Organoids can highly vary in size, shape, and appearance \citep{domenech2023tellu}, and be difficult to distinguish from dust and debris present in the culture \citep{matthews2022organoid,keles2022scalable}.
Finally, the high number of objects to analyze per image poses a big hurdle to a human prone to distraction and fatigue \citep{haja2023towards}.
Manual annotation of this data, which is still state-of-the-art \citep{costa2021drug, wu2022transcriptomics} is, therefore, error- and bias-prone,  which introduces noise in the labels. %, which calls for proper quantification of label uncertainty.
% First, the inconsistency in the assessments made by different individuals (raters) when evaluating the same image (inter-rater) can permit the detection of biases and different expertise levels.
% Second, the variability in the assessments made by one rater when evaluating the same image multiple times (intra-rater) can permit the detection of errors and ambiguities associated with the complexity of the task.
% In most cases of study, the single organoids are manually counted, which is not only time-consuming but also very susceptible to human errors and biases. Additionally, automatic quantification of organoids can greatly assist researchers in determining their size, another important measure of regenerative capacity.
%
However, evaluating learning algorithms for organoid detection, involves comparing predicted outcomes to those manual annotations or '\eac{GT}' during training and testing.
%inter- and intra-rater uncertainties introduce noise in the labels.
As shown in our previous work, deep learning algorithms can outperform even highly-trained human annotators ~\citep{kofler2021we}.
In complex real-life datasets, understanding the shortcomings that label uncertainty creates in the '\eac{GT}' is, therefore, pivotal before training and benchmarking \eac{DL} models.
Moreover, quantifying the label noise by assessing the  intra- and inter-rater reliability is crucial to interpret similarity metrics between model predictions and reference annotations ~\citep{kofler2023approaching}.
% While the need for automated processes that allow organoid quantification is apparent, one must also pay close attention when developing automated models for organoid detection. It is important to consider the aspect of human biases and errors, not only during manual procedures but also when developing and using automatic methods, i.e.
% For this reason, we also avoid using the term \eac{GT} but rather discuss \textit{'annotations'} to account for the label noise potentially introduced by the human annotator.
%For this reason, we collected three sets of annotations, which we name \textit{label sets}, for a part of our data and attempted not only to compare the performance of our benchmark model on each of these but also with respect to each other.

\begin{figure*}
    \centering
    \includegraphics[width=1.0\textwidth]{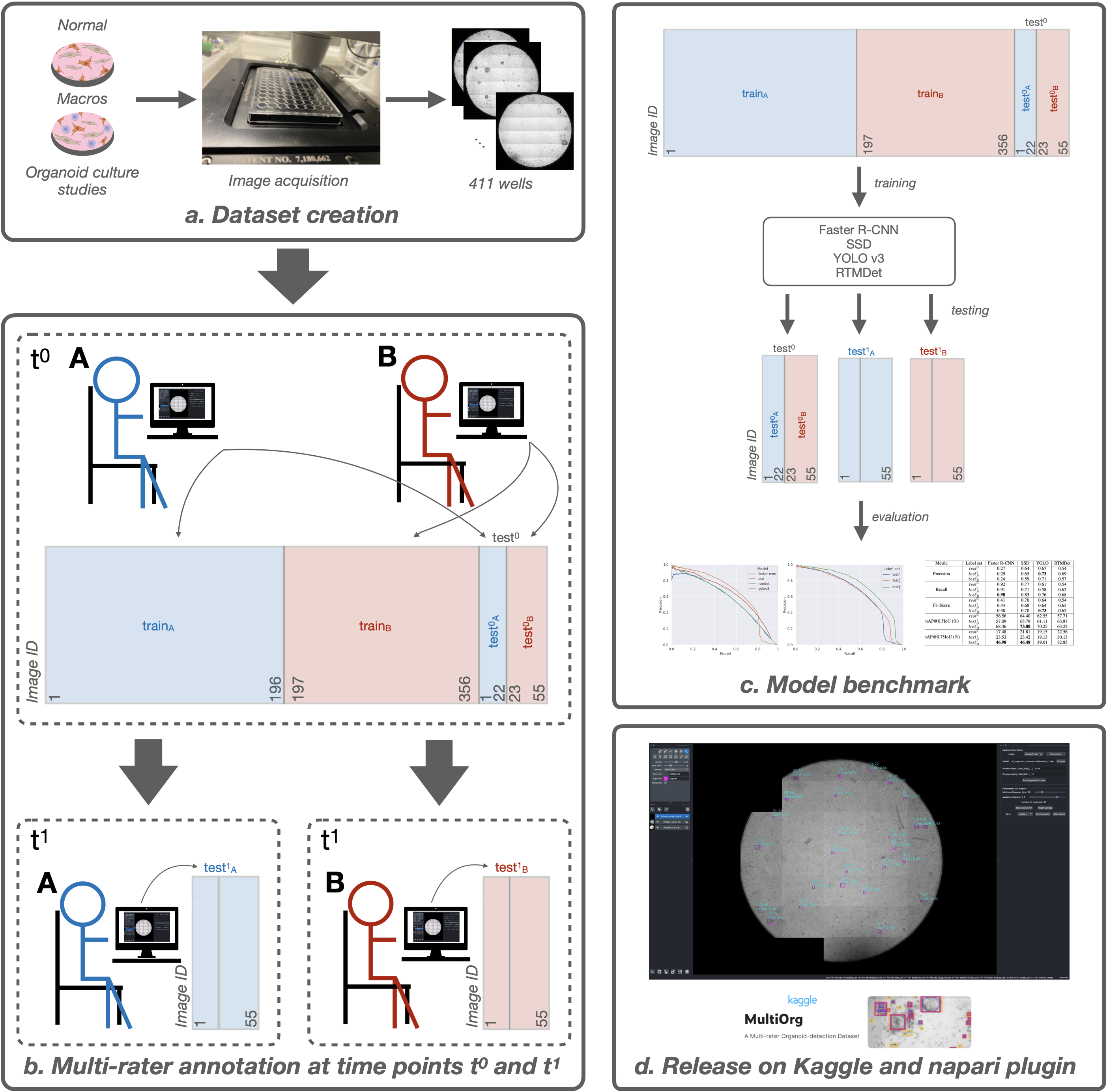}
    \caption{
    \emph{MultiOrg} workflow. a) Dataset creation, b) Multi-rater annotation at time points $t^0$ and $t^1$, c) Model benchmark, and d) Release on Kaggle and napari plugin
    }
    \label{fig:graph_abs}
\end{figure*}

\begin{figure*}
    \centering
    \includegraphics[width=1.0\textwidth]{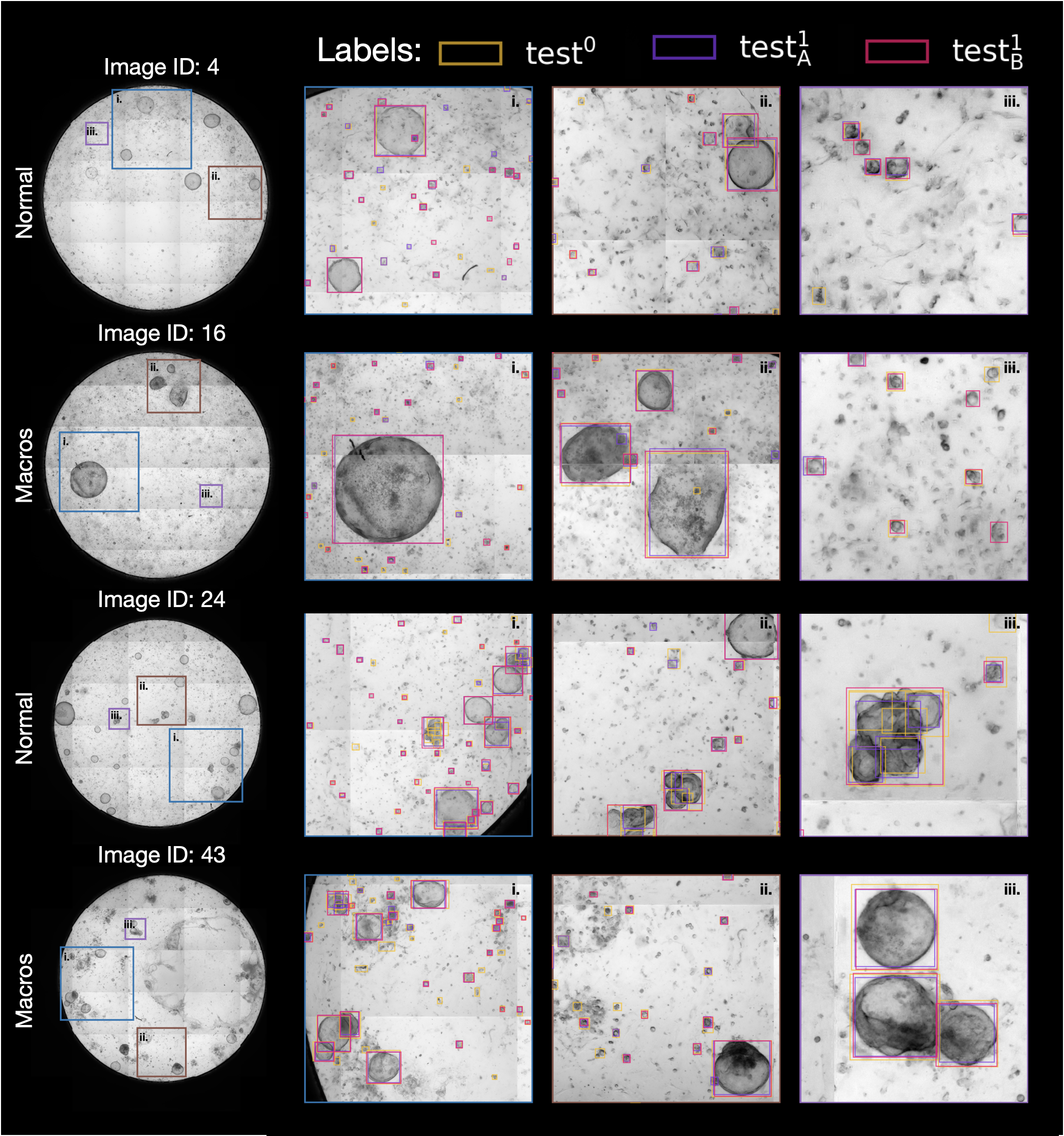}
    \caption{
    Multiple label sets in \emph{MultiOrg}.
    Full test image (left) and crops of areas A, B, and C overlaid with $test^0$, $test^1_A$ and $test^1_B$ (right). 
    The square crops are of sizes 1800, 1200, and 500 px. % and squares of 1800, 1200, and 500 pixels respectively. 
% Images 4 and 24 (respectively 16 and 43) are of the study type 'Normal' (resp. 'Macros').
$test^0$ in images 4 and 16 (respectively 24 and 43) originates from Annotator A (resp. B).
'Macros' are typically noisier, as the cultures initially contain more cells (\Cref{sec:biological_experimental_setup}).%, making it difficult to distinguish between small organoids and debris.
We observe a reduction in the number of annotations at time $t^1$, as the annotators do not consider some small organoids that were annotated at $t^0$.
In image 24, Annotator B annotates clumps of organoids as one large object at $t^1$.
The large structure in image 43 is an experimental matrigel artifact. % (scarring of the matrigel).
The image-wise intra-rater Recall scores are 0.776, 0.532, 0.667 and 0.503 for images 4, 16, 24, and 43, respectively (with $test^0$ as \eac{GT}).
    % 
    % In Image 16, we see an example of how Annotator A does not consider some of the smaller organoids at time point $t^1$ for the Macros case, which results in a low Recall of 0.532 for this image.
    % Also, Annotator B appears to not consider overlapping organoids at time point $t^1$, which can be seen for Image 24, and leave out some smaller organoids which they had labeled at time point $t^0$ (Image 43), which can explain the low Intra-rater Recall observed in \Cref{fig:intra-inter_rater} for Normal and Macros experiments of labeled by Annotator B (Intra-rater Recall for Image 24 is 0.667 and for Image 43 is 0.503).
    }
    \label{fig:enter-label}
\end{figure*}

% CONTRIBUTION
In this work, see \Cref{fig:graph_abs}, we release \emph{MultiOrg}, a large multi-rater 2d microscopy imaging dataset of lung organoids for benchmarking object detection methods.
 % and to further the study of uncertainty estimation methods.
The dataset comprises more than 400 images of an entire microscopy plate well and more than 60,000 annotated organoids, deriving from different biological study setups, with two types of organoids growing under varying conditions.
% Two expert annotators annotated the organoids in the dataset by fitting them within bounding boxes.
% To our knowledge, this is the second largest organoid dataset to date to be made freely available to the community \cite{bremer2022goat}.
Most importantly, we introduce three unique label sets derived from the two annotators at different times, allowing for the quantification of label noise (see Fig.~\ref{fig:enter-label}).
Such a dataset can enable the community to explore biases in annotations, investigate the effect these have on model training, and promote the active area of research for uncertainty quantification.
To our knowledge, this is the second largest organoid dataset to date to be made freely available to the community \citep{bremer2022goat}. It is also the first organoid dataset and one of the very few biomedical object-detection datasets to introduce multiple labels \citep{nguyen2022vindr, amgad2022nucls}.
% This is also the second-largest organoid dataset and the largest lung organoid dataset to be made freely available to the community. % and the first of its kind.
%
We benchmarked this dataset by training and testing four widely established \eac{DL} models for object detection tasks using both one-stage and two-stage architectures.
Finally, along with the dataset and model, we release a tool for quantifying lung organoids, enabling users to visualize and correct the detected organoids before extracting useful features for downstream tasks.
This tool solves the bottleneck of manual quantification of lung organoids, enabling high-throughput image analysis for biological studies. 

% The plugin can be installed from Napari, an open-source image analysis tool, allowing users to efficiently run the algorithm, validate the results, and extract useful features.
In summary, the contributions of this work are as follows:
\begin{itemize}
    \item We release an object detection bio-medical dataset of more than 400 microscopy images comprising around 60,000 lung organoids annotated by two expert annotators.
    \item We provide quantification of label uncertainty through a \href{https://www.kaggle.com/datasets/christinabukas/mutliorg}{Kaggle benchmark challenge} that evaluates the submissions on the different test label sets.
    \item We benchmark our dataset on four standard object detection methods and show how performance varies depending on the selected annotations.
    \item We release the best model in a napari plugin,  \href{https://www.napari-hub.org/plugins/napari-organoid-counter}{\textit{napari-organoid-counter}} \citep{christina_bukas_2022_7065206}, which allows users to curate predictions, thus enabling high-throughput analysis.
\end{itemize}

%The Intra-rater (test0 as GT) F1-Score, Precision and Recall for these images are: 
% Image ID:4, F1=0.815, Prec=0.858, Rec=0.776
% Image ID:16, F1=0.633, Prec=0.789, Rec=0.532
% Image ID:24, F1=0.715, Prec=0.771, Rec=0.667
% Image ID:43, F1=0.643, Prec=0.89, Rec=0.503

\section{Related work}
\label{sec:related_work}
%At present, there are few tools and methods used to detect and quantify organoids, either with traditional image analysis methods or \eac{DL} approaches.
% In recent years, organoid models have become valuable tools in biomedical research and significant advancements have been seen in their development. \citet{du2023organoids} provides an overview of application areas of organoids and describes the development of different machine learning and \eac{DL} methods that address the shortcomings of traditional image processing techniques and the importance of the former in organoid observation and analysis.
% Since imaging modalities, mainly brightfield and fluorescent microscopy, are key in studying their growth and development, it goes without saying that the computer vision community has followed suit. 

\citet{kassis2019orgaquant} proposed \emph{OrganoQuant}, a manually-annotated, human-intestinal-organoid dataset of around 14,000 organoids, along with an object detection pipeline based on Faster R-CNN \citet{ren2015faster}, to locate and quantify human intestinal organoids in brightfield images.
Though object detection performance is satisfactory and the quantification process is robust, inference is performed on cropped patches of a well.
Similarly, \citet{matthews2022organoid} released a dataset of brightfield and phase-contrast microscopy images and proposed an image analysis platform, \emph{OrganoID}, based on U-Net \citet{falk2019u}, which segments and tracks different types of organoids.
They trained their model on images of pancreatic cancer organoids and validated it on pancreatic, lung, colon, and adenoid cystic carcinoma organoids. 
This work introduces several types of organoids. However, the dataset is small, including only 66 images featuring 5 to 50 organoids each.
In \citet{haja2023towards}, \textit{OrganelX} platform was released to enable segmentation of murine liver organoids using Mask-RCNN \citet{he2017mask}.
Furthermore, \citet{bian2021deep} introduced a high-throughput image dataset of liver organoids for detection and tracking.
They also propose a novel deep neural network % (DNN) 
architecture to track organoids dynamically and detect them quickly and accurately.
However, here, too, the dataset size is relatively small, with 75 images containing a total of 6,482 organoids.
\citet{bremer2022goat} used a multicentric dataset consisting of 729 images containing 90,210 annotated organoids, including multiple organoid systems like liver, intestine, tumor, and lung, and proposes an organoid annotation tool, \emph{GOAT}, which uses Mask R-CNN \citep{he2017mask}, for unbiased quantification.
The corresponding dataset contains six organoid types, generated in four centers and acquired with five microscopes.
More recently, \citet{domenech2023tellu} proposed an object detection algorithm, based on YOLO v5 \citet{ultralytics2021yolov5}, \emph{Tellu}, to classify and detect intestinal organoids of different types.
The tool also enables automated analysis of intestinal organoid morphology and fast and accurate classification of organoids.

\emph{MultiOrg} is, therefore, the second largest organoid dataset (see Table~\ref{tab:comparing-datasets}). It is noisier than those introduced above; it is not the densest but contains clumps of organoids and displays an extensive range of sizes, presenting one of the most challenging settings for object detection.
We introduce several label sets on the test set to address this complexity. %This is why we have different annotators and annotations. 
In the dataset, we focused on the detection task only, since it suffices for most practical applications and it is the challenging part from a machine learning point-of-view. Once the detection has been done, the segmentation can be obtained from pre-trained segmentation models (e.g., SAM \citet{kirillov2023segment}).

\begin{table}
    \caption{Overview of the published organoid datasets. We report the organs from which the organoids derive, the number of images, their resolution, the total number of annotated organoids, and the presence of multiple label sets in the dataset. \emph{MultiOrg} is the only one to provide multiple label sets.}
    \label{tab:comparing-datasets}
    \centering
    \resizebox{\columnwidth}{!}{%
        \begin{tabular}{lcccccc}
            \toprule
            \textbf{Dataset} & \textbf{Organ} & \textbf{\# Images} & \textbf{Image Resolution} & \textbf{\# Organoids} & \textbf{Multi-Label} \\
            \midrule
            OrgaQuant \citep{kassis2019orgaquant} & Intestine & 1750 & 300x300, 450x450 & 14,240 & $\times$ \\
            OrganoID \citep{matthews2022organoid} & Pancreas, Lung, Colon, Adenoid & 66 & 512x512 & 5-50/image & $\times$ \\
            \citet{bian2021deep} & Liver & 75 & 7227x7214 & 6,482 & $\times$ \\
            GOAT \citep{bremer2022goat} & Liver, Intestine, Tumor, Lung & 729 & 512x512 & 90,210 & $\times$ \\
            Tellu \citep{domenech2023tellu} & Intestine & 840 & 960x1280 & 23,066 & $\times$ \\
            \emph{MultiOrg} (ours) & Lung & 411 & 6390x5724 & 63,042 & $\checkmark$ \\
            \bottomrule
        \end{tabular}
    }
\end{table}

None of the above-mentioned datasets related to the study of organoids in computer vision offer more than one set of labels.
Nevertheless, comparing multiple annotations in \eac{DL} is not new. 
Various previous initiatives have publicly released multi-rater biomedical datasets for image segmentation \citep{armato2011lung,styner20083d,almazroa2017agreement,lesjak2018novel,mehta2022qu,bran2024qubiq} and classification \citep{orlando2020refuge,sivaswamy2015comprehensive,aung2015automatic}.
Fewer are available, though, for object detection.
To our knowledge, two medical imaging datasets are currently available \citep{nguyen2022vindr,amgad2022nucls}.
The \emph{VinDr-CXR} dataset consists of 18k chest X-ray images annotated with bounding boxes by three radiologists for the presence of 28 lung diseases \citep{nguyen2022vindr}. 
The \emph{NuCLS} dataset provides 97,000 annotations by 32 raters of nuclei from breast cancer pathology images \citep{amgad2022nucls}.
Since labeling uncertainty in object detection is as common as in other image analysis tasks, we hope our dataset will help mitigate the gap in multi-rater detection datasets and contribute to advancing models embracing label variability.

\section{Dataset}
\label{dataset}

\subsection{Dataset creation}
\emph{MultiOrg} consists of 411 bright-field microscopy images representing entire wells of lung organoids derived from murine cells and collected from 26 different studies.
Each study can belong to one of two different types, either 'Normal' or 'Macros' (\Cref{fig:graph_abs}(a) and \Cref{sec:biological_experimental_setup}). % and the images were annotated by two expert annotators. 
During image acquisition, each 3d plate well was imaged in two-dimensional (2d) layers, each divided into smaller tiles (\Cref{{sec:image_acquisition}}).
% Per well, 24 tiles and 12 stacks were acquired.
Individual tiles were then stitched together to form one stack of images.
% We observed that most object detection methods can successfully detect organoids that lie on the borders of two or more patches, even if the stitching mechanism is imperfect. 
% Therefore, in our setup, we decided to work with the stitched images rather than the individual patches, thus tackling the problem of having organoids in multiple patches counted more than once and skipping a preprocessing step without loss in accuracy. 
Since organoids are spherical structures, we applied maximum projection to merge this stack into a single plane, thereby reducing the annotation effort to one image per well, later estimating the organoid volumes from their 2d projection. 

% To train and evaluate our dataset on object detection models (see \Cref{sec:model}), 

For dataset annotation, all organoids present in the images should be fitted by a bounding box.
The annotation process was carried out as in \citet{kastlmeier2023cytokine} by using the initial release, \textit{v.0.1.0}, of the \emph{napari-organoid-counter} tool \citep{christina_bukas_2022_7065206} to generate pseudo labels as a starting point for both annotators with a fixed set of parameters (see \Cref{sec:annotation_procedure}). 
The dataset was initially annotated at time point $t^0$ by two annotators (see \Cref{fig:graph_abs}(b)), namely Annotator A (53\% of the images) and Annotator B (47\% of the images).
% all images of the dataset, both train and test, were annotated once, by Annotator A and the remaining by Annotator B.
The images were then split into train and test sets stratified by annotators and study type. % (see \ref{tab:train-test-numbers} and \Cref{tab:data-overview} for details).
The training set then consists of 356 images derived from 25 studies, and the evaluation was performed on the remaining 55 images from 7 studies as a held-out test set (details in \Cref{tab:data-overview}).
% Annotations were performed at two different time points by two annotators, namely Annotator A and Annotator B (see \Cref{fig:graph_abs}).
% At the first time point, $t^0$, all images of the dataset, both train and test, were annotated once, 53\% by Annotator A and the remaining by Annotator B.
% More specifically, Annotator A annotated 15 experiments, consisting of 218 images out of which 189 images belong to the Normal and 29 images to the Macros setup.
% Annotator B annotated 11 experiments at time point $t^0$, consisting of 193 images out of which 155 belong to the Normal and 38 images belong to the Macros setup. 
% As \Cref{fig:graph_abs}.b shows, out of the 218 images labeled by Annotator A, 196 images were used for training and the remaining 22 for testing.
% Similarly, out of the 193 images labeled by Annotator B, 160 images were used for training and 33 for testing. 
At time point $t^1$, Annotators A and B reannotated all test images, blinded to their initial labels (\Cref{fig:enter-label}).
Annotator A used the same setup, whereas Annotator B changed their setup (computer mouse and monitor) between $t^0$ and $t^1$.
The number of organoids annotated by each annotator, in each study type, for the train and test sets are provided in \Cref{tab:train-test-numbers} and
statistics on bounding box sizes in \Cref{fig:area_annot_labelset} and \Cref{tab:area_annot_labelset}.

We, therefore, provide three label sets for the images of our test set.
The annotations produced at time point $t^0$ are denoted $test^0$ and can further be split into subsets $test^0_A$ and $test^0_B$ since images 1-22 were annotated by A and 23-55 by B (see \Cref{fig:graph_abs}).
Additionally, label sets $test^1_A$ (respectively $test^1_B$) refer to the re annotation from annotators A (resp. B), at time point $t^1$ on all 55 images of the test set. 

% In the training dataset, the average bounding box size of organoids belonging to the Normal and Macros setups annotated by Annotator A are \SI{11,733}{\micro\meter\squared} and \SI{6,676}{\micro\meter\squared} respectively, while the overall average size bounding box size of organoids annotated by Annotator A is \SI{11,329}{\micro\meter\squared}. For images annotated by Annotator B, the average bounding box size of organoids for Normal, Macros and the combined setups are \SI{11,633}{\micro\meter\squared}, \SI{10,867}{\micro\meter\squared}, and \SI{11,571}{\micro\meter\squared} respectively. 

\begin{table}[ht]
    \caption{
    Overview of the label sets (\textit{train}, \textit{test}$^0$, \textit{test}$^1_A$, and \textit{test}$^1_B$).
    Number of images and organoid labels stratified by study type (for all) and annotator (only relevant for \textit{train} and \textit{test}$^0$).
    All \textit{test} label sets refer to the same images.
    We see a reduction in the number of labels between $t^0$ and $t^1$.
    }
    \label{tab:train-test-numbers}
    \centering
    \resizebox{0.9\columnwidth}{!}{%
        \begin{tabular}{lcccccc}
            \toprule
            \textbf{Study Type} & \multicolumn{2}{c}{Normal} & \multicolumn{2}{c}{Macros} & \multicolumn{2}{c}{Combined} \\
            \cmidrule(r){2-3} \cmidrule(r){4-5} \cmidrule(r){6-7}
            & \# Images & \# Organoids & \# Images & \# Organoids & \# Images & \# Organoids \\
            \midrule
            \multicolumn{7}{c}{\textbf{Train set}} \\
            \midrule
            \textit{train}$_A$ & 181 & 30,710 & 15 & 2,669 & 196 & 33,379 \\
            \textit{train}$_B$ & 135 & 20,263 & 25 & 1,781 & 160 & 22,044 \\
            \textbf{Total} & \textbf{316} & \textbf{50,973} & \textbf{40} & \textbf{4,450} & \textbf{356} & \textbf{55,423} \\
            \midrule
            \multicolumn{7}{c}{\textbf{Test set}} \\
            \midrule
            \textit{test}$^0_A$ & 8 & 1,145 & 14 & 1,865 & 22 & 3,010 \\
            \textit{test}$^0_B$ & 20 & 3,020 & 13 & 1,493 & 33 & 4,513 \\
            \textbf{Total (Label set \textit{test}$^0$)} & \textbf{28} & \textbf{4,165} & \textbf{27} & \textbf{3,358} & \textbf{55} & \textbf{7,523} \\
            \midrule
            \textbf{Label set \textit{test}$^1_A$} & \textbf{28} & \textbf{2,748} & \textbf{27} & \textbf{1,981} & \textbf{55} & \textbf{4,729} \\
            \midrule
            \textbf{Label set \textit{test}$^1_B$} & \textbf{28} & \textbf{2,655} & \textbf{27} & \textbf{2,301} & \textbf{55} & \textbf{4,956} \\
            \bottomrule
        \end{tabular}
    }
\end{table}

\subsection{Object detection metrics }
\label{sec:object_detection_tasl_eval_metrics}

We compare the multiple label sets and assess the quality of model predictions using several evaluation metrics. 
% that take these aspects into account. 
% When comparing the different label sets or model predictions to each other, one needs to be defined as the '\eac{GT}'.
\eac{TPs}, \eac{FPs}, and \eac{FNs} are computed for each image, for a given \eac{IOU} threshold, using one of the label sets as the '\eac{GT}'.
Their total numbers are then aggregated on the entire test set.
For comparing the three available label sets, we compute Precision and Recall at an \eac{IOU} of 0.5 and the F1-score.  
While Precision measures the percentage of correct predictions against a considered true label, Recall (i.e., sensitivity) measures the proportion of true positive predictions identified correctly.
For evaluating model performance, we use the \eac{P-R} curves as the primary tool.
It consists of Precision and Recall values at different model confidence thresholds, at a fixed \eac{IOU} threshold (here we use 0.5 unless specified otherwise).
We also report \eac{mAP}, by integrating precision across Recall levels from 0 to 1.
We compute them using the standard library \citep{electronics10030279} which follows the PASCAL VOC challenge technique for interpolating points on the curve \citep{pascal-voc-2012}.
\subsection{Multi-rater analysis}
\label{sec:quantification_label_noise}
% Understanding and quantifying human errors and biases is an essential aspect of supervised machine learning across various computer vision applications, from image classification to segmentation to object detection. 
% Only by understanding the discrepancies in what is often accepted as \textit{"ground truth"} can one truly estimate the performance of machine learning models, which are trained and evaluated based on the fact that these data are error-free.
% As mentioned in the previous section, the proposed dataset contains three label sets for the images of the test set.
% These can estimate the variance in the annotations and help discern label noise introduced by the annotators. 
% We compute the Inter- and Intra-rater reliability scores between and within our two annotators to quantify this variance and assess their consistency.
% Intra-rater analysis is the comparison of annotations by the same annotator at different time points, whereas Inter-rater analysis provides a comparison of annotations of the same images but annotated by a different annotator at the same or different time points.

% Understanding and quantifying human errors and biases in labels is crucial for supervised machine learning in computer vision applications, as it helps to estimate model performance by addressing shortcomings of the \textit{"ground truth"}.
% As discussed, the proposed dataset contains three label sets for the images of the test set.
We compute inter- and intra-rater uncertainties to quantify annotation variance and assess the consistency of the two raters over time and against each other.
Inter-rater scores assess the inconsistency in the assessments made by different raters when evaluating the same image and can permit the detection of biases and different expertise levels.
The variability in assessments made by one rater when evaluating the same image multiple times (intra-rater) can permit the detection of errors and ambiguities associated with the complexity of the task.

\begin{figure}[ht]
    \centering
\includegraphics[width=1.0\columnwidth]{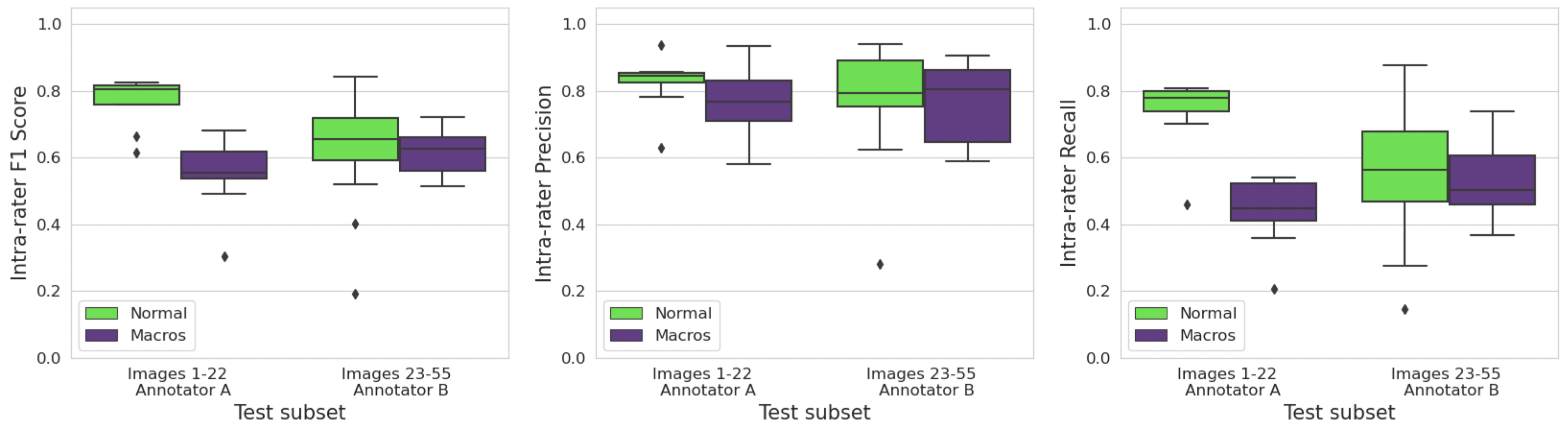}
\includegraphics[width=1.0\columnwidth]{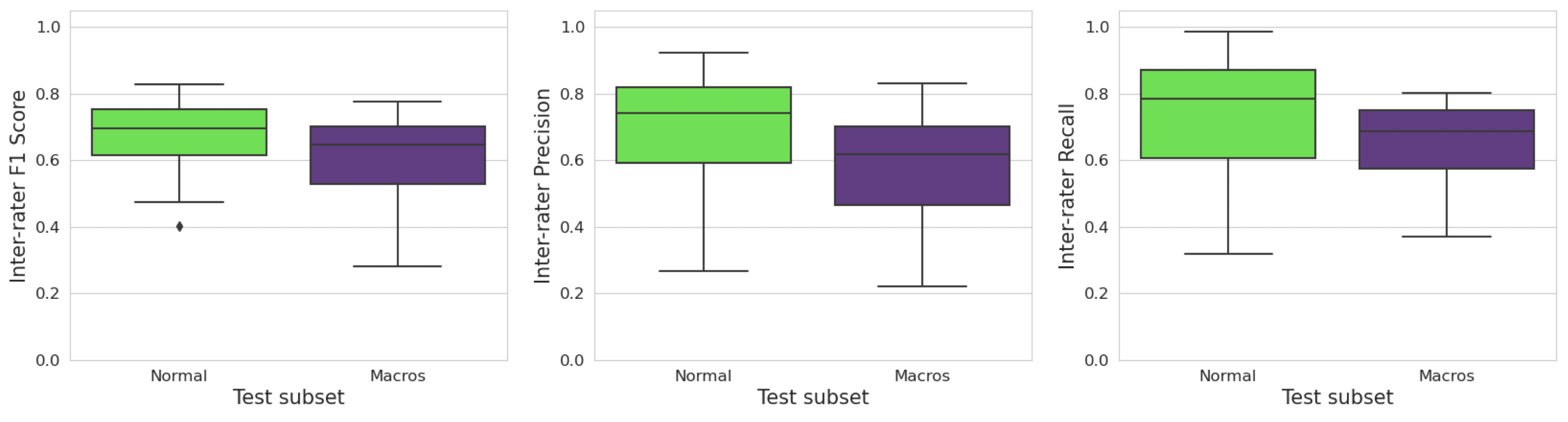}
    \caption{ Multi-rater scores.
    \textbf{Top}: Intra-rater F1-score (left), Precision (middle), and Recall (right), where $test^0$ is considered the \eac{GT}, for both annotators and according to study type.
    Annotator A appears more consistent on 'Normal' images (higher scores), and annotation of 'Macros' seems more challenging (with lower scores).
    Both annotators show an overall higher Precision and lower Recall, indicating that $test^0$ has many more annotations which are treated here as \eac{FNs}.
    \textbf{Bottom}: Inter-rater F1-score (left), Precision (middle), and Recall (right) on the test set between $test^1_A$ and $test^1_B$, where $test^1_A$ is considered the \eac{GT}, split according to study type.
    Raters agree more on 'Normal' images, indicating that the annotation of 'Macros' images is more challenging.
    Individual differences are generally lower than in-between raters (lower inter-rater than intra-rater scores). 
% (the spurious $t^0$ annotations become here \eac{FNs})
    }
    \label{fig:intra_inter_rater}
\end{figure}

\Cref{fig:intra_inter_rater} shows intra-rater scores, with label set $test^0$ used as the \eac{GT} (note that switching the choice of \eac{GT} does not impact the F1-score and switchs Recall and Precision).
For Annotator A, we compare the annotations for test images 1-22, i.e., $test^0_A$ with the corresponding subset of $test^1_A$, while for Annotator B we compare annotations of images 23-55, i.e., $test^0_B$ with the corresponding subset of $test^1_B$.
We find that Annotator A is more consistent over time, especially for the 'Normal' studies, which is in line with the reported change of annotation setup of Annotator B.
Additionally, annotation of 'Macros' seems more challenging than 'Normal' images.
We also find a reduction in the number of annotations at time point $t^1$, already reported in \Cref{tab:train-test-numbers}.
The qualitative inspection of \Cref{fig:enter-label} suggests that some pseudo labels used as a starting point for all manual annotations were not removed in $test^0$, probably indicating improvement of the annotations over time.
Further comparison of the pseudo labels to the label sets can be seen in \Cref{tab:pseudo}.
We also observe that Annotator B, unlike Annotator A, annotated overlapping clumps of organoids differently at $t^1$, which could be the source of the higher reported inconsistency.
% More precisely, in Image 16, we see an example of how
% Annotator A does not consider some of the smaller organoids at time point $t^1$ for the Macros case, which results in a low Recall for this image.
% Also, Annotator B appears to not consider overlapping organoids at time point $t^1$, which can be seen for Image 24, and leave out some smaller
% organoids which they had labeled at time point $t^0$ (Image 43), in line with a low Intra-rater Recall (Intra-rater Recall for Image 24 is 0.667 and for Image 43 is 0.503).
 % observed in ?? for Normal and Macros experiments of labeled by Annotator B (Intra-rater Recall for Image 24 is 0.667 and for Image 43 is 0.503).

% Another observation in line with the above can be drawn by studying \Cref{tab:multi-rater-analysis}, where higher F1-scores are found between annotations from time point $t^1$ (mean on the two subsets is 0.610 and 0.710 for $test^1_A$ vs. $test^1_B$, compared to 0.586 and 0.561 for $test^0$ vs. the subsets at timepoint $t^1$).
%Naturally, intra-rater scores are higher than inter-rater scores (annotators agree more with themselves at a new time point than with others)

Inter-rater scores can be computed for the entire test set between $test^1_A$ and $test^1_B$, as shown in \Cref{fig:intra_inter_rater}.
We again observe that the annotation of 'Macros' is generally more challenging, and those images consistently appear noisier in \Cref{fig:enter-label}.
% one can see that images deriving from the Macros experiments are typically noisier and it can be difficult to distinguish between small organoids and debris.
%
In \Cref{fig:inter_rater_over_time}(bottom), we also display the inter-rater scores on the image subsets 1-22 and 23-55, as well as across $t^0$ and $t^1$.
This corroborates the larger evolution of Annotator B between $t^0$ and $t^1$ and indicates a convergence of the two annotation styles.
% \Cref{fig:intra-inter_rater} provides an overview of intra- and inter-rater scores, where label set $test^0$ is used as the \eac{GT} to compute Recall, Precision, and F1-score at 0.5 \eac{IOU}.
% In \Cref{fig:inter_rater_over_time} we use the inter-rater score to observe how the labels from one annotator change over time by comparing them to an independent third label set. 
% For example, the Inter-rate F1-score between $test^1_A-test^0_B$ and $test^1_A-test^1_B$ shows that the annotation style of Annotator B changes between time points $t^0$ and $t^1$.
\Cref{tab:multi-rater-analysis}, \Cref{tab:multi-rater-analysis-normal}, and \Cref{tab:multi-rater-analysis-macro}, provide all statistics for these multi-rater scores.

%Therefore, Inter-rater scores can be computed for the entire test set between \textit{Label set t1\_A} and \textit{Label set t1\_B}.

% Therefore, Inter-rater scores can be computed for the entire test set between \textit{Label set t0\_AB} and images originally annotated by Annotator A in \textit{Label set t1\_A} in addition to images originally annotated by Annotator B in \textit{Label set t1\_B}.
% \textit{Label set t1\_A} and \textit{Label set t1\_B}.
%To compute the Intra-rater scores for Annotator A, we compare the annotations from \textit{Label set t0\_AB} that were annotated by Annotator A, termed as \textit{t0\_AB\_A}, to the corresponding annotations in \textit{Label set t1\_A}, termed as \textit{Label set t1\_A\_A}.
%Similarly, to compute the Intra-rater scores for Annotator B, we compare the annotations from \textit{Label set t0\_AB} that were annotated by Annotator B, termed as \textit{t0\_AB\_B}, to the corresponding annotations in \textit{Label set t1\_B}, termed as \textit{Label set t1\_B\_B}.

\subsection{Dataset availability}
\label{dataset_availability}
We make \emph{MultiOrg} available to the community. 
All images are public on \href{https://www.kaggle.com/datasets/christinabukas/mutliorg}{Kaggle}, together with label sets $train$ and $test^0$, to ensure that the steps presented in \Cref{sec:model} can be reproduced.
The label sets $test^1_A$ and $test^1_B$ can be queried by participating in the \href{https://www.kaggle.com/competitions/multi-org-challenge}{\emph{MultiOrg} challenge},
where our leaderboard returns the average of \eac{mAP} on $test^1_A$ and $test^1_B$.
We invite scientists to participate, to promote research in the field of uncertainty estimation. 

%[TODO] Discuss whether we want one big availability section where everything goes [model, napari tool, image data + annotations]

%[TODO mention licensing]
%\begin{itemize}
%    \item images
%    \item labels
%\end{itemize}

\section{Model Benchmarking}
\label{sec:model}
% To show the  evaluate the performance of object detection methods on our dataset and with the goal in mind of enabling future users to build upon these for furthering the development of noise-aware models,
We benchmark four standard object-detection \eac{DL} models on \emph{MultiOrg}:
\begin{itemize}
    \item \textit{Faster R-CNN} \citep{ren2015faster}
    \item \eac{SSD} \citep{liu2016ssd}
    \item \eac{YOLOv3} \citep{redmon2018yolov3}
    \item \eac{RTMDet} \citep{lyu2022rtmdet}.
\end{itemize}

All trained models can be found on \href{https://zenodo.org/records/11258022?token=eyJhbGciOiJIUzUxMiJ9.eyJpZCI6ImQ5ZTAxYjQ1LTRiZGYtNDA5OC04N2MzLWM0ZThkYzhhZDE5MSIsImRhdGEiOnt9LCJyYW5kb20iOiJmNTU1Y2EzNjVkYzg1MDc0MjdiMDkyNjk2MzdkNmFhZSJ9.WFGdz7HYplboUWQPxu72CNNBIx3_CdgaIaix76ukIzqPq2eIsnAFrp0IdpbC-Q6EKGzpaajdnxX4iXzcqfSIyA}{zenodo}, and code and documentation to reproduce the training is available on \href{https://www.kaggle.com/datasets/christinabukas/mutliorg}{Kaggle} \footnote{In addition to the tested methods, we initially implemented DETR with standard hyper-parameters for our dataset, but the training was quite unstable, and the performance much worse than other models. We therefore decided not to report the scores in the manuscript.
Furthermore, MedSAM \citep{MedSAM} and Cellpose \citep{stringer2021cellpose}, segmentation-based approaches, did not work well out-of-the box. Since they would also require segmentation labels for fine-tuning to be useful we did not include these methods for benchmarking.}.

%Our proposed method is inspired by OrgaQuant \citep{kassis2019orgaquant}.
%We implemented a two-stage object detection model, using the \eac{FRCNN} architecture \citep{ren2015faster}, which uses ResNet50 as the backbone.
%In the first stage, a \eac{RPN} is designed which generates potential region proposals.
% \eac{RPN} is essentially a neural network with an attention mechanism, which helps in finding the potential regions where objects are present.
%The region proposals from the first stage are then used as input to the second stage and object classes and bounding boxes are predicted.

\subsection{Training and Testing}
\label{sec:model-training}
%To train our object detection model, we used a \eac{DL} pipeline called Quicksetup-ai \citep{author_year}, which is based on PyTorch Lightning and the Hydra framework.
For training, the images in the training set were split into patches of 512x512 px, resulting in a total of 20,011 patches.
Bounding boxes extending beyond the borders of the patches were omitted in the \eac{GT} since these organoids can be captured by a sliding window approach at inference, resulting in 44,418 bounding box labels for training.
For training and validation, we used the \textit{mmdetection} \citep{mmdetection} toolbox, with the original configuration for each model adapted such that the input and parameters for all models is the same (details in \Cref{sec:appendix-model}).

During testing, sliding window inference was performed on the full images of the test set. 
We slide over each image twice with different window sizes and down-sampling factors to detect both small and large organoids.
% If the window size is too small, then larger organoids may not be detected, whereas if the window size is too big, then smaller organoids can be missed.
We empirically choose the following parameters: window size set to 512 and 2048 px, while the down-sampling factor is set to two and eight, respectively, with a window overlap of 0.5 and \eac{NMS} for post-processing with a threshold of 0.5.
For each model, we choose the checkpoint with the highest \eac{mAP} on $test^0$, thus using this label set for validation during training. 
We report those, along with training and inference times in \Cref{tab:train_test_params}.
%The model confidence threshold, \eac{BST}, can be tuned during inference for each model.
%If the confidence score of a prediction is less than the \eac{BST} value then that box is removed.
%Based on the literature, we record the best-performing metric at a \eac{BST} of 0.5.

%\eac{FRCNN} is jointly trained with a combination of four losses: two cross-entropy losses, one binary for classifying the regions proposed by the RPN and one for classifying the final bounding boxes, and two smooth \textit{l1} regression losses, again for the outputs of the RPN and Fast R-CNN network.
%\eac{FRCNN} is jointly trained with a combination of four losses: \(loss\_objectness\), \(loss\_rpn\_box\_reg\), \(loss\_cls\) and \(loss\_box\_reg\). \(loss\_objectness\) is the loss related to the proposed prediction of an object by \eac{RPN} and \(loss\_rpn\_box\_reg\) is the loss related to proposed bounding boxes by \eac{RPN} (log loss object or not object).\(loss\_box\_reg\) is the loss of the prediction of bounding box coordinates (regression loss → smooth L1), and  \(loss\_cls\) is the loss of prediction of object classes in bounding boxes. 

%\input{body/04_baseline/042_inference}
\subsection{Benchmark models evaluation}
\label{sec:model_res}
\begin{figure}[htbp!]
    \centering
    \includegraphics[width=0.9\textwidth]{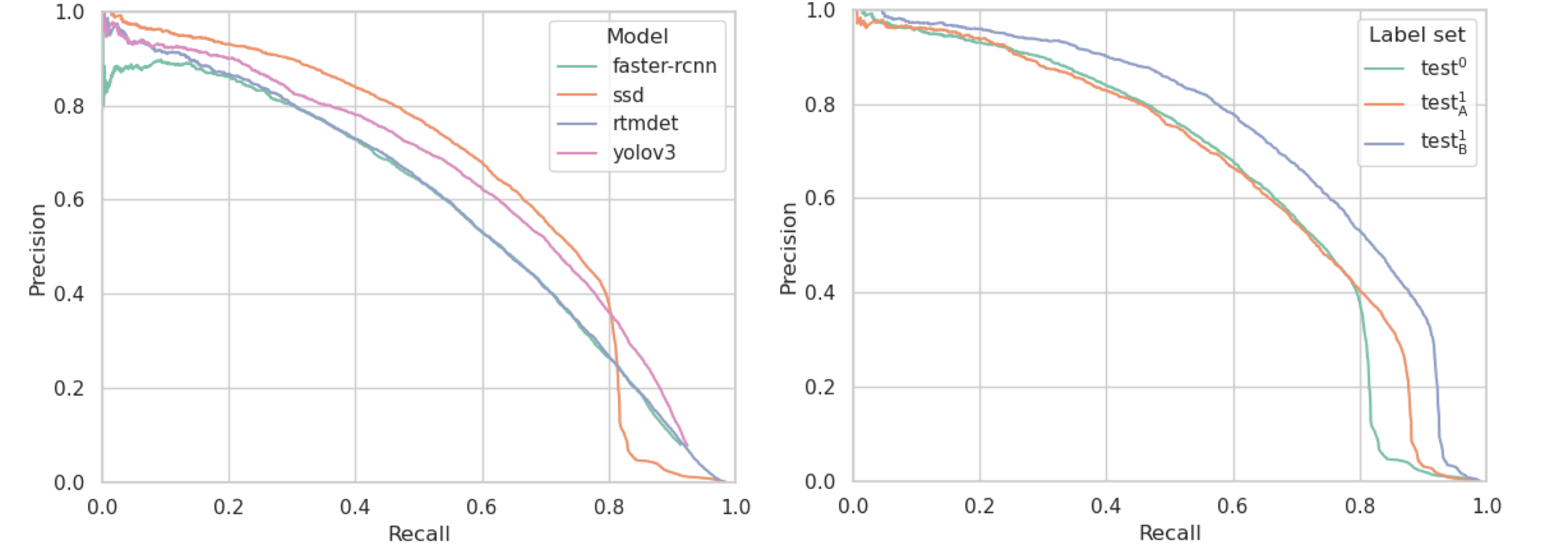}
    \caption{Model Benchmark. \eac{P-R} curves using $test^0$ as the \eac{GT} for all models (left) and using all three label sets for \eac{SSD} (right).
    We observe that overall the SSD model predictions are more in agreement with the annotations and have a better trade-off between precision and recall.
    Although the model was trained and validated with labels from $t^0$ it is more in agreement with annotations from timepoint  $t^1$.}
    \label{fig:pr-curve}
\end{figure}

We evaluate the model performance on $test^0$,  $test^1_A$ and  $test^1_B$.
\Cref{fig:pr-curve} shows \eac{P-R} curves for the different models on label set $test^0$, as well as the curves for all three label sets on the best performing model, \eac{SSD}.
Notably, though the model was trained and validated on labels created at $t^0$, the best \eac{P-R} curve is obtained for $test^1_B$, indicating that the trained model is more in agreement with these labels.
This suggests that the higher label noise present in $test^0$, as assumed in \Cref{sec:quantification_label_noise}, was not picked up by the model during training, illustrating once more the resilience of \eac{DL} to label noise \citep{rolnick2017deep}.
\Cref{tab:model_benchmark_res} presents further evaluation metrics and confirms that \eac{SSD} is the best-performing model overall,  while at the standard model confidence threshold of 0.5 \eac{YOLOv3} performs equally well if not better on some label sets.
Interestingly, different models exhibit very different Precision-Recall trade-offs at 0.5 model confidence.
% We observe that \eac{SSD} performs better overall, as can also be confirmed by the \eac{P-R} curves in \Cref{fig:pr-curve}, while at the standard \eac{BST} of 0.5 \eac{YOLOv3} performs equally well if not better on some label sets.

\begin{table}[ht]
    \caption{
    Benchmark metrics on the three label sets. Precision, Recall, and F1-score are reported at 0.5 \eac{IOU} threshold and model confidence. \eac{mAP} is reported at 0.5 and 0.75 IoU threshold.
    The models exhibit different Precision-Recall tradeoffs. The performance of \eac{SSD} is overall better when considering \eac{mAP}, while at the standard model confidence threshold of 0.5 \eac{YOLOv3} performs equally well if not better on some label sets.
    }
    \label{tab:model_benchmark_res}
    \centering
    \resizebox{0.8\columnwidth}{!}{%
        \begin{tabular}{cccccc}
            \toprule
            \textbf{Metric} & \textbf{Label set} & \textbf{Faster R-CNN} & \textbf{SSD} & \textbf{YOLOv3} & \textbf{RTMDet} \\
            \midrule
            \multirow{4}{*}{Precision} & $test^{0}$ & 0.23 & 0.61 & \textbf{0.73} & 0.64 \\
            & $test^{1}_{A}$ & 0.16 & 0.44 & 0.58 & 0.54 \\
            & $test^{1}_{B}$ & 0.18 & 0.50 & 0.67 & 0.56 \\
            \cline{2-6}
            & \textbf{mean} & \textbf{0.19} & \textbf{0.52} & \textbf{0.66} & \textbf{0.58} \\
            \midrule
            \multirow{4}{*}{Recall} & $test^{0}$ & 0.84 & 0.67 & 0.48 & 0.51 \\
            & $test^{1}_{A}$ & 0.92 & 0.78 & 0.62 & 0.69 \\
            & $test^{1}_{B}$ & \textbf{0.97} & 0.83 & 0.67 & 0.68 \\
            \cline{2-6}
            & \textbf{mean} & \textbf{0.91} & \textbf{0.76} & \textbf{0.59} & \textbf{0.63} \\
            \midrule
            \multirow{4}{*}{F1-score} & $test^{0}$ & 0.36 & 0.64 & 0.58 & 0.57 \\
            & $test^{1}_{A}$ & 0.27 & 0.57 & 0.60 & 0.61 \\
            & $test^{1}_{B}$ & 0.30 & 0.62 & \textbf{0.67} & 0.62 \\
            \cline{2-6}
            & \textbf{mean} & \textbf{0.31} & \textbf{0.61} & \textbf{0.62} & \textbf{0.60} \\
            \midrule
            \multirow{4}{*}{mAP@0.5IoU (\%)} & $test^{0}$ & 56.56 & 64.40 & 62.55 & 57.71 \\
            & $test^{1}_{A}$ & 57.09 & 65.79 & 61.11 & 63.87 \\
            & $test^{1}_{B}$ & 68.36 & \textbf{73.88} & 70.25 & 63.23 \\
            \cline{2-6}
            & \textbf{mean} & \textbf{60.67} & \textbf{68.09} & \textbf{64.64} & \textbf{61.60}  \\
            \midrule
            \multirow{4}{*}{mAP@0.75IoU (\%)} & $test^{0}$ & 17.48 & 21.81 & 19.15 & 22.56 \\
            & $test^{1}_{A}$ & 23.53 & 23.42 & 19.13 & 30.13 \\
            & $test^{1}_{B}$ & \textbf{46.98} & \textbf{46.48} & 39.01 & 32.85 \\
            \cline{2-6}
            & \textbf{mean} & \textbf{29.33} & \textbf{30.57} & \textbf{25.76} &  \textbf{28.51} \\
            \bottomrule
        \end{tabular}
    }
\end{table}

\subsection{Napari plugin} % for Napari enables researchers to automatize organoid quantification}
\label{sec:napari-plugin}
% Napari is an open-source image analysis tool used to visualize and analyze multi-dimensional images and annotations \citep{Ahlers_napari_a_multi-dimensional}.
As described in \Cref{sec:annotation_procedure}, \emph{MultiOrg} was created using the open-source image analysis tool \emph{Napari} \citep{Ahlers_napari_a_multi-dimensional}, together with the initial release, \textit{v.0.1.0}, of the \href{https://www.napari-hub.org/plugins/napari-organoid-counter}{\textit{napari-organoid-counter}} plugin \citep{christina_bukas_2022_7065206}.
% , to generate pseudo labels and manually curate than they then manually curate the dataset proposed in this work, resulting in more than 60,000 annotated organoids (see \Cref{sec:annotation_procedure}).
%
In this work, we release a new version of the plugin, \textit{v.0.2.2}, using the model from our benchmark with the better trade-off between performance and inference time as the backbone (see \Cref{tab:train_test_params}), i.e. \eac{YOLOv3}, along with added functionalities.
For example, the model confidence threshold
% is added as a feature at run time and 
can now be adjusted at run time by the user, % depending on the user's choice, 
depending on whether for the task at hand, a higher Recall or Precision is most practical. %important.
% Further highlights of the latest version of the \href{https://www.napari-hub.org/plugins/napari-organoid-counter}{\textit{napari-organoid-counter}}
Plugin details can be found in \Cref{sec:app-napari-plugin}.
Code and tutorials for installation are distributed through the
 \href{https://www.napari-hub.org/plugins/napari-organoid-counter}{napari hub} 
%\href{https://www.youtube.com/watch?v=dQw4w9WgXcQ}{napari hub}.

\section{Discussion}
\label{sec:discussion}
%-GT discussion: while the average dataset comes with a single annotation that they declare as GT (even though strictly speaking GT is often unachievable in biomedical data) we embrace this complexity and present a label-noise-aware benchmark with multiple annotations [TODO cite peak GT], that comes with quantification of label noise

% \paragraph{Contribution:} 
In this work, we release \emph{MultiOrg}, a large multi-rater dataset for organoid detection in 2d microscopy images.
Our dataset consists of more than 60,000 annotated lung organoids, labeled by two expert annotators.
As even expert annotators can disagree on what constitutes an organoid in those images while also being susceptible to human error and biases, we provide three label sets for the test data, enabling quantification of label uncertainty on a multi- and single-annotator level.
Additionally, we have carefully included diversity in our dataset through several study setups and cell lines to ensure good generalization. We performed preliminary tests of the selected model on different lung cell types not present in the dataset, from both human and mouse organoids, and seen that it generalizes quite well. We also tested it successfully on colon organoids and speculate that it can be used for all organoids with similar shape and size.
%While the research stemming from the study of organoids is steadily growing \citet{kim2020human}, the demand for automatic quantification tools becomes ever more imminent.
%To the best of our knowledge, such an object detection dataset, with multiple annotations in the field of microscopy, has never before been released. 
This dataset is, therefore, uniquely situated between the fields of microscopy and uncertainty quantification.
We invite researchers to use it and assist us in studying label noise in challenging real-life biomedical settings,
and we believe that future models should, as much as possible, refrain from being trained without considering these aspects.
%Though only single labels exist for the training set, new methods can still take into account the three label sets available for the test set.
%For this purpose we have released our test data only on Kaggle, ....
We additionally publish a benchmark for organoid object detection, provide all models in \href{https://zenodo.org/records/11258022?token=eyJhbGciOiJIUzUxMiJ9.eyJpZCI6ImQ5ZTAxYjQ1LTRiZGYtNDA5OC04N2MzLWM0ZThkYzhhZDE5MSIsImRhdGEiOnt9LCJyYW5kb20iOiJmNTU1Y2EzNjVkYzg1MDc0MjdiMDkyNjk2MzdkNmFhZSJ9.WFGdz7HYplboUWQPxu72CNNBIx3_CdgaIaix76ukIzqPq2eIsnAFrp0IdpbC-Q6EKGzpaajdnxX4iXzcqfSIyA}{zenodo} and the best one as a \href{https://www.napari-hub.org/plugins/napari-organoid-counter}{Napari plugin}, thus enabling scientists potentially use them on their own data.
% to analyze and experiment on our dataset, as well as potentially find uses for their own data.

%TODO contribution + implications for the real-world

% \paragraph{Limitations \& Future Work:}

This work offers a valuable dataset that can be leveraged to advance both object detection methods and uncertainty quantification techniques.
The current setting does not, however, permit the incorporation of several label sets in the training loop.
Furthermore, it is important to note that although two organoid types are present in the images, % (those grown from AEC2, and those including macrophages),
they have not been annotated as different classes. %there is currently no distinction between them in the current version of the dataset labels.
Treating this task as a multi-class detection problem may boost the overall performance of the object detection task \citep{zhang2022glipv2} and would provide added value to the biologists.
However, while providing several label sets or multi-class labels on the training data could be beneficial, it represents substantial manual work. 
% Further work could include providing multiple label sets on the training data, that could be incorporated in the training loop, by exploring various techniques, such as ensemble learning \citep{mirikharaji2021d}.
 % While multiple labels are currently provided only for the test split and cannot be used in the training loop, researchers can still leverage the dataset for significant insights and improvements. 
% While this work provides a valuable dataset that can be used by researchers to further develop both object detection methods and uncertainty quantification techniques alike, one needs to remember that multiple labels are only provided for the test split and, therefore, cannot be incorporated into the training loop.
% An additional future release could incorporate additional label sets also for the training data, which could assist model training, by exploring various techniques, such as ensemble learning \citep{mirikharaji2021d}. 
%Moreover, the dataset could be extended, to include label sets from more annotators, or even, by the existing annotators to produce further label sets for comparison by curating the predictions of the different benchmark models.
It would also be interesting to observe how the label sets change by using \eac{DL} models as a baseline for annotation rather than the pseudo labels. % used in the current version.
Despite these limitations, we are confident that the release of the \emph{MultiOrg} dataset offers several invaluable contributions to the machine learning community.

\section*{Acknowledgements}
This work was funded by the German Center for Lung Research (DZL) and from the Deutsche Forschungsgemeinschaft (DFG, German Research Foundation– 512453064) as well as from the Stiftung Atemweg. We would like to thank Kathrin Federl for her excellent technical assistance during the creation of the dataset.
We would also like to thank Isra Mekki and Francesco Campi for reviewing the code for reproducing the benchmark, and Theresa Willem for her help with the Ethics Statement.

% \clearpage
{
    \small
    \bibliographystyle{ieeenat_fullname}
    \bibliography{references}
}

%%%%%%%%%%%%%%%%%%%%%%%%%%%%%%%%%%%%%%%%%%%%%%%%%%%%%%%%%%%%
\section*{Checklist}

% %%% BEGIN INSTRUCTIONS %%%
% The checklist follows the references.  Please
% read the checklist guidelines carefully for information on how to answer these
% questions.  For each question, change the default \answerTODO{} to \answerYes{},
% \answerNo{}, or \answerNA{}.  You are strongly encouraged to include a {\bf
% justification to your answer}, either by referencing the appropriate section of
% your paper or providing a brief inline description.  For example:
% \begin{itemize}
%   \item Did you include the license to the code and datasets? \answerYes{See Section~\ref{gen_inst}.}
%   \item Did you include the license to the code and datasets? \answerNo{The code and the data are proprietary.}
%   \item Did you include the license to the code and datasets? \answerNA{}
% \end{itemize}
% Please do not modify the questions and only use the provided macros for your
% answers.  Note that the Checklist section does not count towards the page
% limit.  In your paper, please delete this instructions block and only keep the
% Checklist section heading above along with the questions/answers below.
% %%% END INSTRUCTIONS %%%

\begin{enumerate}

\item For all authors...
\begin{enumerate}
  \item Do the main claims made in the abstract and introduction accurately reflect the paper's contributions and scope?
    \answerYes{}
  \item Did you describe the limitations of your work?
    \answerYes{See \Cref{sec:discussion}}
  \item Did you discuss any potential negative societal impacts of your work?
    \answerYes{See \cref{sec:ethics}}
  \item Have you read the ethics review guidelines and ensured that your paper conforms to them?
    \answerYes{See \cref{sec:ethics}} 
\end{enumerate}

\item If you are including theoretical results...
\begin{enumerate}
  \item Did you state the full set of assumptions of all theoretical results?
    \answerNA{}
	\item Did you include complete proofs of all theoretical results?
    \answerNA{}
\end{enumerate}

\item If you ran experiments (e.g. for benchmarks)...
\begin{enumerate}
  \item Did you include the code, data, and instructions needed to reproduce the main experimental results (either in the supplemental material or as a URL)?
    \answerYes{Links for all assets released are provided in \Cref{sec:availability}}
  \item Did you specify all the training details (e.g., data splits, hyperparameters, how they were chosen)?
    \answerYes{See \Cref{tab:data-overview}, \Cref{sec:model-training}, and \Cref{sec:appendix-model}}
	\item Did you report error bars (e.g., with respect to the random seed after running experiments multiple times)?
    \answerNo{}
	\item Did you include the total amount of compute and the type of resources used (e.g., type of GPUs, internal cluster, or cloud provider)?
    \answerYes{See \Cref{sec:model-training} and \Cref{tab:train_test_params}}
\end{enumerate}

\item If you are using existing assets (e.g., code, data, models) or curating/releasing new assets...
\begin{enumerate}
  \item If your work uses existing assets, did you cite the creators?
    \answerYes{}
  \item Did you mention the license of the assets?
    \answerNA{}
  \item Did you include any new assets either in the supplemental material or as a URL?
    \answerYes{See \cref{sec:availability}, or alternatively \cref{dataset_availability}, \cref{sec:model}, and \cref{sec:napari-plugin}}
  \item Did you discuss whether and how consent was obtained from people whose data you're using/curating?
    \answerYes{}
  \item Did you discuss whether the data you are using/curating contains personally identifiable information or offensive content?
    \answerYes{}
\end{enumerate}

\item If you used crowdsourcing or conducted research with human subjects...
\begin{enumerate}
  \item Did you include the full text of instructions given to participants and screenshots, if applicable?
    \answerNA{}
  \item Did you describe any potential participant risks, with links to Institutional Review Board (IRB) approvals, if applicable?
    \answerNA{}
  \item Did you include the estimated hourly wage paid to participants and the total amount spent on participant compensation?
    \answerNA{}
\end{enumerate}

\end{enumerate}

%%%%%%%%%%%%%%%%%%%%%%%%%%%%%%%%%%%%%%%%%%%%%%%%%%%%%%%%%%%%

\newpage
\appendix
\section{Appendix}
\counterwithin{figure}{section}
\counterwithin{table}{section}
\subsection{Data Description}

\subsubsection{Biological study setup}
\label{sec:biological_experimental_setup}
Two different study setups were used to create our dataset.
In the first, we used isolated murine distal epithelial cells, enriched for alveolar epithelial type II cells (AEC2), which have the stem-cell function of differentiating into other cell types, hence serving as progenitor cells in the lung \citep{barkauskas2013type}.
In addition to the AEC2, we used a murine fibroblast cell line as mesenchymal support cells.
Images containing organoids deriving from these biological cultures, are henceforth mentioned as belonging to the 'Normal' setup.

As a second study setup, we added a cell type, namely macrophages, to the organoid culture, which we henceforth refer to as the 'Macros' setup.  
Isolation of cells and culturing of organoids were performed as previously described in \citet{lehmann2020chronic}.
Organoids were seeded as duplicates in 96-well imaging plates with a glass bottom (see \Cref{fig:graph_abs}).

\subsubsection{Image acquisition}
\label{sec:image_acquisition}
The plates used for culturing and imaging were \emph{Falcon®} 96-well Black/Clear Flat Bottom TC-treated Imaging Microplates.
%Imaging took place at different time points of culture (specify time points).
The brightfield images were acquired with a Life Cell Imaging Microscope \emph{(LifeCellImagerObserver.Z1)} at a 5x objective.
During the acquisition of the images, each well was divided into tiles and stacks to capture the 3d growth of organoids.
Per well, 24 tiles and 10-15 stacks were acquired. %, measuring in \(\mu m\) were set.
Individual tiles were stitched together to form one single image per plate.
We observed that most object detection methods can successfully detect organoids on the borders of two or more patches, even if the stitching mechanism is imperfect. 
Therefore, in our setup, we decided to work with the stitched images rather than the individual patches. %, thus tackling the problem of having organoids in multiple patches counted more than once and skipping a preprocessing step without loss in accuracy. 

Maximum projections were generated by the \emph{Zen 2 Blue} software by \emph{Carl Zeiss Microscopy GmbH} to process the image stacks into one plain 2D image, such that each pixel in the final image derives from the slice in the stack which is most in focus at that location.
Since organoid structures are relatively spherical, one can easily approximate their area using a 2D projection. 
% Furthermore, the thickness of the solution used was small, not often allowing for multiple organoids on the z-axis, except for very small structures, allowing to convert the images into 2D without losing significant information.
% Due to the thickness of the solution, the culture is effectively 3d, and small organoids can thus occasionally overlap in the projected image.
% Small organoids can overlap in the projection on the z-axis, except for very small structures, allowing to convert the images into 2D without losing significant information.
Additionally, labeling images in 2D greatly speeds up the annotation procedure.
Images were exported in the \emph{CZI} file format.

The resulting 2D images have varying sizes, between 5719 and 6240 pixels in the \textit{x} and 5551 and 6940 pixels in the \textit{y} axis, respectively. Each pixel in the image is equivalent to 1.29 \(\mu m\) in each axis.
At this point, the images were examined, and eight plates were dropped, either due to lower image quality or because the organoid formation did not work well, resulting in noisy images.
% , with an overlarge number of organoids present, making it hard to visualize.
The latter was mainly observed in the Macros study setup, which resulted in fewer data from this setup in our final dataset.
Finally, the imaged wells were randomly selected by plates and study setups to be annotated using our annotation tool of choice (see \Cref{sec:annotation_procedure}).

\subsubsection{Annotation Procedure}
\label{sec:annotation_procedure}
The annotation process was carried out by running the initial release, \textit{v.0.1.0}, of the \emph{napari-organoid-counter} tool \citep{christina_bukas_2022_7065206}, a plugin developed for \emph{Napari} \citep{Ahlers_napari_a_multi-dimensional}. %, an open-source Python-based tool for image visualization, annotation, and analysis.
%Two individual expert annotators labeled the dataset at two different time-points \textit{t0} and \textit{t1}.
The tool parameters were set to a down-sampling of one, minimal diameter of 30 µm and sigma of three.
After running the \emph{napari-organoid-counter} with these parameters, all detected organoids were examined.
All spherical structures consisting of visibly more than one cell and measuring more than 30 µm, were recognized as organoids.
The wrongfully created box was manually deleted if the counter detected a \eac{FP}.
If the counter detected the organoid size or exact location incorrectly, the box was manually moved or adjusted according to the correct size and location.
If the counter detected accumulations of organoids as one single object, the box was deleted and correctly sized boxes for the single organoids were created.
If the counter did not detect an organoid, a box according to the organoid's size was manually created.
% After checking the whole well and correcting the boxes, the number of detected organoids was updated, and the file was saved by generating an Excel sheet with the amount and exact measurements of the detected boxes surrounding the organoid.
% With the number of detected organoids, the colony-forming efficiency of the progenitor cells was calculated using the input number of progenitor cells and used as reference size.
% The measurements of the single boxes were used to calculate the diameter of the organoids and displayed in graphs.

%As discussed in \Cref{dataset}, at $t^0$ all images of the dataset, both train and test, were annotated once, 53\% by Annotator A and the remaining by Annotator B.
%At time point $t^1$, Annotators A and B each annotated once more the images belonging to our test set.
%We therefore procured three sets of annotations for the images of our test set, which we hereafter call \textit{label sets}.
%The annotations produced at time point $t^0$ correspond to label set $test^0$, and can further be split into subsets $test^0_A$ and $test^0_B$, since images 1-22 were annotated by A and 23-55 by B (see \Cref{fig:graph_abs}).
%Additionally, label sets $test^1_A$ and $test^1_B$ refer to the annotations of annotators A and B respectively at time point $t^1$ on all 55 images of the test set. 

\FloatBarrier

\subsubsection{Data preparation for release}
After all the data was collected and annotated, all images were converted from the proprietary CZI to the open TIFF format. Additionally, all studies were renamed to ensure consistency and all images and annotation information for each image were anonymized. 

\begin{table}[ht]
    \caption{
    A detailed overview of the dataset. The training set consists of 356 images derived from 25 studies, and the test set consists of 55 images from 7 studies.
    }
    \label{tab:data-overview}
    \centering
    \resizebox{0.8\columnwidth}{!}{%
        \begin{tabular}{cccccc}
            \toprule
            \textbf{Study Setup} & \textbf{Plate Name} & \textbf{Number of Wells} & \textbf{Image IDs} & \textbf{Data split} & \textbf{Annotator} \\
            \midrule
            Normal & Plate\_11 & 13 & 1-13 & train & A \\
            Normal & Plate\_13 & 1 & 14 & train & A \\
            Macros & Plate\_13 & 6 & 15-20 & train & A \\
            Normal & Plate\_19 & 6 & 21-26 & train & A \\
            Normal & Plate\_20 & 20 & 27-46 & train & A \\
            Normal & Plate\_26 & 34 & 47-80 & train & A \\
            Normal & Plate\_29 & 26 & 81-106 & train & A \\
            Normal & Plate\_3 & 5 & 107-111 & train & A \\
            Normal & Plate\_32 & 19 & 112-130 & train & A \\
            Normal & Plate\_33 & 16 & 131-146 & train & A \\
            Normal & Plate\_34 & 16 & 147-162 & train & A \\
            Macros & Plate\_6 & 9 & 163-171 & train & A \\
            Normal & Plate\_8 & 10 & 172-181 & train & A \\
            Normal & Plate\_9 & 15 & 182-196 & train & A \\
            Normal & Plate\_16 & 17 & 197-213 & train & B \\
            Macros & Plate\_16 & 9 & 214-222 & train & B \\
            Normal & Plate\_17 & 18 & 223-240 & train & B \\
            Macros & Plate\_17 & 4 & 241-244 & train & B \\
            Normal & Plate\_18 & 34 & 245-278 & train & B \\
            Macros & Plate\_18 & 6 & 279-284 & train & B \\
            Normal & Plate\_24 & 10 & 285-294 & train & B \\
            Macros & Plate\_25 & 6 & 295-300 & train & B \\
            Normal & Plate\_36 & 15 & 301-315 & train & B \\
            Normal & Plate\_39 & 17 & 316-332 & train & B \\
            Normal & Plate\_40 & 24 & 333-356 & train & B \\
            Normal & Plate\_37 & 6 & 1-6 & test & A \\
            Normal & Plate\_4 & 2 & 7-8 & test & A \\
            Macros & Plate\_4 & 14 & 9-22 & test & A \\
            Normal & Plate\_15 & 12 & 23-34 & test & B \\
            Normal & Plate\_31 & 8 & 35-42 & test & B \\
            Macros & Plate\_15 & 7 & 43-49 & test & B \\
            Macros & Plate\_23 & 6 & 50-55 & test & B \\
            \bottomrule
        \end{tabular}
    }
\end{table}

\begin{figure}[htbp!]
    \centering
    \includegraphics[width=0.7\columnwidth]{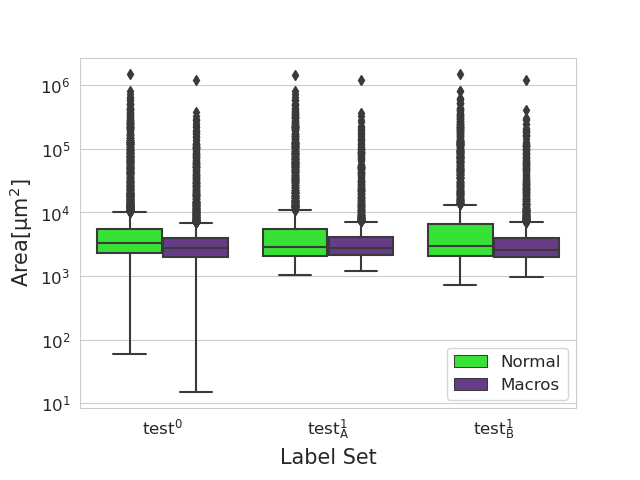}
    \caption{Bounding box sizes. Box plots of the bounding box areas in $test^0$, $test^1_A$ and $test^1_B$,  stratified by study type, on a logarithmic scale.}
    % The crossmark represents the mean area for the specific boxplot and the number on top of the crossmark represents the total number of organoids (outliers) having an area greater than Q3.
% The mean bounding box size (\textpm std.) of organoids in label set $test^0$ for 'Normal', 'Macros' and combined images annotated by Annotator A are \SI{11,887}{\micro\meter\squared}(\textpm \SI{51,000}{\micro\meter\squared}), \SI{4,891}{\micro\meter\squared}(\textpm \SI{31,918}{\micro\meter\squared}) and \SI{7,553}{\micro\meter\squared}(\textpm \SI{40,400}{\micro\meter\squared}) respectively. For Annotator B, it is \SI{17,064}{\micro\meter\squared}(\textpm \SI{57,976}{\micro\meter\squared}), \SI{9,654}{\micro\meter\squared}(\textpm \SI{25,272}{\micro\meter\squared}) and \SI{14,612}{\micro\meter\squared}(\textpm \SI{49,726}{\micro\meter\squared}) respectively.
%     The mean bounding box size of organoids in Label set $test^1_A$ for Normal, Macros and combined images are \SI{19,413}{\micro\meter\squared}(\textpm \SI{64,855}{\micro\meter\squared}), \SI{8,714}{\micro\meter\squared}(\textpm \SI{36,128}{\micro\meter\squared}) and \SI{14,931}{\micro\meter\squared}(\textpm \SI{14,931}{\micro\meter\squared}) respectively.
%     The average bounding box size of organoids in Label set $test^1_B$ for Normal, Macros and combined images are \SI{21,315}{\micro\meter\squared}(\textpm \SI{68,956}{\micro\meter\squared}), \SI{8,188}{\micro\meter\squared}(\textpm \SI{35,042}{\micro\meter\squared}), and \SI{15,220}{\micro\meter\squared}(\textpm \SI{56,215}{\micro\meter\squared}) respectively.}
    \label{fig:area_annot_labelset}
\end{figure}

\begin{table}[h!]
    \centering
    \caption{
    Bounding box sizes. Mean and standard deviation of the bounding box areas (in \SI{}{\micro\meter\squared}) in $test^0_A$, $test^0_B$, $test^1_A$, and $test^1_B$, stratified by study type and combined.
    }
    \label{tab:area_annot_labelset}
    \resizebox{0.7\columnwidth}{!}{%
        \begin{tabular}{cccc}
            \toprule
            \diagbox{Label Subset}{Study Type} & Normal & Macros & Combined \\
            \midrule
            $test^0_A$ & 11,887(\(\pm\) 51,000) & 4,891(\(\pm\) 31,918) & 7,553(\(\pm\) 40,400) \\
            $test^0_B$ & 17,064(\(\pm\) 57,976) & 9,654(\(\pm\) 25,272) & 14,612(\(\pm\) 49,726) \\
            $test^1_A$ & 19,413(\(\pm\) 64,855) & 8,714(\(\pm\) 36,128) & 14,931(\(\pm\) 14,931) \\
            $test^1_B$ & 21,315(\(\pm\) 68,956) & 8,188(\(\pm\) 35,042) & 15,220(\(\pm\) 56,215) \\
            \bottomrule
        \end{tabular}
    }
\end{table}

\begin{comment}
\begin{table}[h!]
    \centering
    {
        \begin{tabular}{| *{10}{c|} }
            \hline
                \diagbox{Subset}{Study Type}
                & 'Normal'
                & 'Macros'
                & combined \\
            \hline
             $test^0_A$  & 11,887(\textpm 51,000) & 4,891(\textpm 31,918) & 7,553(\textpm 40,400) \\
            \hline
             $test^0_B$ & 17,064(\textpm 57,976) & 9,654(\textpm 25,272) & 14,612(\textpm 49,726) \\
            \hline
           $test^1_A$ & 19,413(\textpm 64,855) & 8,714(\textpm 36,128) & 14,931(\textpm 14,931) \\
            \hline
           $test^1_B$ & 21,315(\textpm 68,956) & 8,188(\textpm 35,042) & 15,220(\textpm 56,215) \\
            \hline
        \end{tabular}
        }
    \caption{Bounding box sizes. Mean size (in \SI{}{\micro\meter\squared}) and standard deviation of the bounding box areas in $test^0_A$, $test^0_B$, $test^1_A$ and $test^1_B$, stratified by study type and combined.}
    \label{tab:area_annot_labelset}
\end{table}
\end{comment}

\FloatBarrier
\subsubsection{Multi-rater analysis}

\begin{table}[ht]
    \centering
    \caption{
    Statistics of the multi-rater analysis. The intra-rater scores for Annotator A are calculated for $test^0_A$ (considered as \eac{GT}) and the corresponding subset of $test^1_A$ for images 1-22. Similarly, for Annotator B, the intra-rater score is computed between $test^0_B$ (considered as \eac{GT}) and the corresponding subset of $test^1_B$ for images 23-55. Inter-rater scores for images 1-22 are calculated between $test^0_A$ (considered as \eac{GT}) and the corresponding subset of $test^1_B$, as well as subsets of $test^1_A$ (considered as \eac{GT}) and $test^1_B$. Similarly, for images 23-55, the inter-rater scores are computed between $test^0_B$ (considered as \eac{GT}) and the corresponding subset of $test^1_A$, as well as subsets of $test^1_A$ (considered as \eac{GT}) and $test^1_B$.
    }
    \label{tab:multi-rater-analysis}
    \resizebox{0.9\columnwidth}{!}{%
        \begin{tabular}{cccccccccc}
            \toprule
            \multirow{3}{*}{} & \multicolumn{9}{c}{\textbf{Images 1-22}} \\
            \cmidrule(lr){2-10}
             & \multicolumn{3}{c}{\textbf{Intra-rater Annotator A}}
             & \multicolumn{3}{c}{\textbf{Inter-rater $test^0_A$ vs. $test^1_B$}}
             & \multicolumn{3}{c}{\textbf{Inter-rater $test^1_A$ vs. $test^1_B$}} \\
            \cmidrule(lr){2-10}
             & F1-Score & Precision & Recall & F1-Score & Precision & Recall & F1-Score & Precision & Recall \\
            \midrule
            \textbf{Median} & 0.624 & 0.806 & 0.516 & 0.586 & 0.813 & 0.469 & 0.610 & 0.689 & 0.604 \\
            \textbf{Mean} & 0.635 & 0.787 & 0.551 & 0.577 & 0.782 & 0.473 & 0.605 & 0.677 & 0.595 \\
            \textbf{Std} & 0.134 & 0.094 & 0.173 & 0.090 & 0.091 & 0.116 & 0.110 & 0.175 & 0.159 \\
            \textbf{min} & 0.305 & 0.581 & 0.207 & 0.397 & 0.569 & 0.258 & 0.360 & 0.276 & 0.320 \\
            \textbf{25\%} & 0.545 & 0.726 & 0.439 & 0.523 & 0.731 & 0.414 & 0.547 & 0.593 & 0.481 \\
            \textbf{50\%} & 0.624 & 0.801 & 0.516 & 0.586 & 0.813 & 0.470 & 0.610 & 0.689 & 0.604 \\
            \textbf{75\%} & 0.766 & 0.848 & 0.740 & 0.649 & 0.854 & 0.547 & 0.690 & 0.822 & 0.734 \\
            \textbf{max} & 0.826 & 0.809 & 0.939 & 0.754 & 0.889 & 0.692 & 0.767 & 0.922 & 0.848 \\
            \midrule
            \multirow{3}{*}{} & \multicolumn{9}{c}{\textbf{Images 23-55}} \\
            \cmidrule(lr){2-10}
             & \multicolumn{3}{c}{\textbf{Intra-rater Annotator B}}
             & \multicolumn{3}{c}{\textbf{Inter-rater $test^0_B$ vs. $test^1_A$}}
             & \multicolumn{3}{c}{\textbf{Inter-rater $test^1_A$ vs. $test^1_B$}} \\
            \cmidrule(lr){2-10}
             & F1-Score & Precision & Recall & F1-Score & Precision & Recall & F1-Score & Precision & Recall \\
            \midrule
            \textbf{Median} & 0.643 & 0.800 & 0.546 & 0.561 & 0.725 & 0.426 & 0.710 & 0.667 & 0.761 \\
            \textbf{Mean} & 0.632 & 0.778 & 0.551 & 0.522 & 0.752 & 0.414 & 0.655 & 0.608 & 0.761 \\
            \textbf{Std} & 0.127 & 0.136 & 0.155 & 0.140 & 0.159 & 0.145 & 0.145 & 0.188 & 0.133 \\
            \textbf{min} & 0.193 & 0.282 & 0.147 & 0.219 & 0.348 & 0.148 & 0.281 & 0.222 & 0.381 \\
            \textbf{25\%} & 0.568 & 0.729 & 0.461 & 0.434 & 0.650 & 0.324 & 0.559 & 0.480 & 0.704 \\
            \textbf{50\%} & 0.643 & 0.800 & 0.546 & 0.561 & 0.725 & 0.426 & 0.710 & 0.667 & 0.761 \\
            \textbf{75\%} & 0.703 & 0.885 & 0.667 & 0.611 & 0.909 & 0.503 & 0.760 & 0.750 & 0.835 \\
            \textbf{max} & 0.844 & 0.940 & 0.878 & 0.807 & 1.000 & 0.732 & 0.828 & 0.831 & 0.988 \\
            \bottomrule
        \end{tabular}
    }
\end{table}

\begin{table}[ht]
    \centering
    \caption{
    Statistics of the multi-rater analysis for 'Normal' images. The intra-rater scores for Annotator A are calculated for $test^0_A$ (considered as \eac{GT}) and the corresponding subset of $test^1_A$ for images 9-22. Similarly, for Annotator B, the intra-rater score is computed between $test^0_B$ (considered as \eac{GT}) and the corresponding subset of $test^1_B$ for images 43-55. Inter-rater scores for images 9-22 are calculated between $test^0_A$ (considered as \eac{GT}) and the corresponding subset of $test^1_B$, as well as subsets of $test^1_A$ (considered as \eac{GT}) and $test^1_B$. Similarly, for images 43-55, the inter-rater scores are computed between $test^0_B$ (considered as \eac{GT}) and the corresponding subset of $test^1_A$, as well as subsets of $test^1_A$ (considered as \eac{GT}) and $test^1_B$.
    }
    \label{tab:multi-rater-analysis-normal}
    \resizebox{0.9\columnwidth}{!}{%
        \begin{tabular}{cccccccccc}
            \toprule
            \multirow{3}{*}{} & \multicolumn{9}{c}{\textbf{Images 1-8}} \\
            \cmidrule(lr){2-10}
             & \multicolumn{3}{c}{\textbf{Intra-rater Annotator A}}
             & \multicolumn{3}{c}{\textbf{Inter-rater $test^0_A$ vs. $test^1_B$}}
             & \multicolumn{3}{c}{\textbf{Inter-rater $test^1_A$ vs. $test^1_B$}} \\
            \cmidrule(lr){2-10}
             & F1-Score & Precision & Recall & F1-Score & Precision & Recall & F1-Score & Precision & Recall \\
            \midrule
            \textbf{Median} & 0.804 & 0.845 & 0.779 & 0.584 & 0.843 & 0.444 & 0.610 & 0.843 & 0.475 \\
            \textbf{Mean} & 0.768 & 0.824 & 0.735 & 0.584 & 0.848 & 0.458 & 0.612 & 0.834 & 0.513 \\
            \textbf{Std} & 0.081 & 0.090 & 0.117 & 0.100 & 0.028 & 0.127 & 0.096 & 0.082 & 0.171 \\
            \textbf{min} & 0.617 & 0.629 & 0.459 & 0.454 & 0.814 & 0.308 & 0.475 & 0.651 & 0.320 \\
            \textbf{25\%} & 0.761 & 0.826 & 0.740 & 0.532 & 0.827 & 0.382 & 0.558 & 0.822 & 0.410 \\
            \textbf{50\%} & 0.804 & 0.845 & 0.779 & 0.584 & 0.843 & 0.444 & 0.610 & 0.843 & 0.475 \\
            \textbf{75\%} & 0.817 & 0.855 & 0.799 & 0.638 & 0.867 & 0.524 & 0.696 & 0.882 & 0.600 \\
            \textbf{max} & 0.826 & 0.939 & 0.809 & 0.754 & 0.889 & 0.692 & 0.737 & 0.922 & 0.848 \\
            \midrule
            \multirow{3}{*}{} & \multicolumn{9}{c}{\textbf{Images 23-42}} \\
            \cmidrule(lr){2-10}
             & \multicolumn{3}{c}{\textbf{Intra-rater Annotator B}}
             & \multicolumn{3}{c}{\textbf{Inter-rater $test^0_B$ vs. $test^1_A$}}
             & \multicolumn{3}{c}{\textbf{Inter-rater $test^1_A$ vs. $test^1_B$}} \\
            \cmidrule(lr){2-10}
             & F1-Score & Precision & Recall & F1-Score & Precision & Recall & F1-Score & Precision & Recall \\
            \midrule
            \textbf{Median} & 0.657 & 0.794 & 0.564 & 0.578 & 0.833 & 0.444 & 0.738 & 0.707 & 0.823 \\
            \textbf{Mean} & 0.645 & 0.784 & 0.564 & 0.559 & 0.819 & 0.435 & 0.696 & 0.632 & 0.823 \\
            \textbf{Std} & 0.155 & 0.150 & 0.179 & 0.139 & 0.161 & 0.134 & 0.117 & 0.166 & 0.104 \\
            \textbf{min} & 0.193 & 0.282 & 0.147 & 0.219 & 0.348 & 0.160 & 0.404 & 0.268 & 0.595 \\
            \textbf{25\%} & 0.593 & 0.753 & 0.469 & 0.486 & 0.744 & 0.356 & 0.639 & 0.532 & 0.758 \\
            \textbf{50\%} & 0.657 & 0.794 & 0.564 & 0.578 & 0.833 & 0.444 & 0.738 & 0.707 & 0.823 \\
            \textbf{75\%} & 0.719 & 0.891 & 0.679 & 0.636 & 0.932 & 0.504 & 0.779 & 0.750 & 0.891 \\
            \textbf{max} & 0.844 & 0.940 & 0.878 & 0.807 & 1.000 & 0.684 & 0.828 & 0.822 & 0.988 \\
            \bottomrule
        \end{tabular}
    }
\end{table}

\begin{table}[ht]
    \centering
    \caption{
    Statistics of the multi-rater analysis for 'Macros' images. The intra-rater scores for Annotator A are calculated for $test^0_A$ (considered as \eac{GT}) and the corresponding subset of $test^1_A$ for images 9-22. Similarly, for Annotator B, the intra-rater score is computed between $test^0_B$ (considered as \eac{GT}) and the corresponding subset of $test^1_B$ for images 43-55. Inter-rater scores for images 9-22 are calculated between $test^0_A$ (considered as \eac{GT}) and the corresponding subset of $test^1_B$, as well as subsets of $test^1_A$ (considered as \eac{GT}) and $test^1_B$. Similarly, for images 43-55, the inter-rater scores are computed between $test^0_B$ (considered as \eac{GT}) and the corresponding subset of $test^1_A$, as well as subsets of $test^1_A$ (considered as \eac{GT}) and $test^1_B$.
    }
    \label{tab:multi-rater-analysis-macro}
    \resizebox{0.9\columnwidth}{!}{%
        \begin{tabular}{cccccccccc}
            \toprule
            \multirow{3}{*}{} & \multicolumn{9}{c}{\textbf{Images 9-22}} \\
            \cmidrule(lr){2-10}
             & \multicolumn{3}{c}{\textbf{Intra-rater Annotator A}}
             & \multicolumn{3}{c}{\textbf{Inter-rater $test^0_A$ vs. $test^1_B$}}
             & \multicolumn{3}{c}{\textbf{Inter-rater $test^1_A$ vs. $test^1_B$}} \\
            \cmidrule(lr){2-10}
             & F1-Score & Precision & Recall & F1-Score & Precision & Recall & F1-Score & Precision & Recall \\
            \midrule
            \textbf{Median} & 0.556 & 0.769 & 0.449 & 0.586 & 0.744 & 0.486 & 0.621 & 0.615 & 0.682 \\
            \textbf{Mean} & 0.560 & 0.766 & 0.446 & 0.573 & 0.744 & 0.482 & 0.600 & 0.587 & 0.643 \\
            \textbf{Std} & 0.091 & 0.092 & 0.090 & 0.087 & 0.093 & 0.113 & 0.120 & 0.148 & 0.136 \\
            \textbf{min} & 0.305 & 0.581 & 0.207 & 0.397 & 0.569 & 0.258 & 0.360 & 0.276 & 0.370 \\
            \textbf{25\%} & 0.539 & 0.712 & 0.411 & 0.522 & 0.667 & 0.439 & 0.546 & 0.512 & 0.517 \\
            \textbf{50\%} & 0.556 & 0.769 & 0.449 & 0.586 & 0.744 & 0.486 & 0.621 & 0.615 & 0.682 \\
            \textbf{75\%} & 0.620 & 0.832 & 0.524 & 0.646 & 0.812 & 0.557 & 0.688 & 0.698 & 0.748 \\
            \textbf{max} & 0.683 & 0.935 & 0.542 & 0.684 & 0.871 & 0.640 & 0.767 & 0.771 & 0.804 \\
            \midrule
            \multirow{3}{*}{} & \multicolumn{9}{c}{\textbf{Images 43-55}} \\
            \cmidrule(lr){2-10}
             & \multicolumn{3}{c}{\textbf{Intra-rater Annotator B}}
             & \multicolumn{3}{c}{\textbf{Inter-rater $test^0_B$ vs. $test^1_A$}}
             & \multicolumn{3}{c}{\textbf{Inter-rater $test^1_A$ vs. $test^1_B$}} \\
            \cmidrule(lr){2-10}
             & F1-Score & Precision & Recall & F1-Score & Precision & Recall & F1-Score & Precision & Recall \\
            \midrule
            \textbf{Median} & 0.626 & 0.806 & 0.503 & 0.491 & 0.659 & 0.373 & 0.648 & 0.655 & 0.704 \\
            \textbf{Mean} & 0.613 & 0.768 & 0.529 & 0.464 & 0.650 & 0.381 & 0.592 & 0.569 & 0.665 \\
            \textbf{Std} & 0.063 & 0.117 & 0.111 & 0.127 & 0.087 & 0.160 & 0.164 & 0.219 & 0.116 \\
            \textbf{min} & 0.515 & 0.590 & 0.368 & 0.226 & 0.476 & 0.148 & 0.281 & 0.222 & 0.116 \\
            \textbf{25\%} & 0.560 & 0.647 & 0.461 & 0.405 & 0.586 & 0.296 & 0.523 & 0.433 & 0.636 \\
            \textbf{50\%} & 0.626 & 0.806 & 0.503 & 0.491 & 0.659 & 0.373 & 0.648 & 0.655 & 0.704 \\
            \textbf{75\%} & 0.662 & 0.862 & 0.606 & 0.561 & 0.705 & 0.435 & 0.727 & 0.700 & 0.752 \\
            \textbf{max} & 0.723 & 0.905 & 0.738 & 0.651 & 0.790 & 0.732 & 0.776 & 0.831 & 0.778 \\
            \bottomrule
        \end{tabular}
    }
\end{table}

\begin{figure}[ht]
    \centering
    \makebox[\textwidth][c]{\includegraphics[width=1.2\columnwidth]{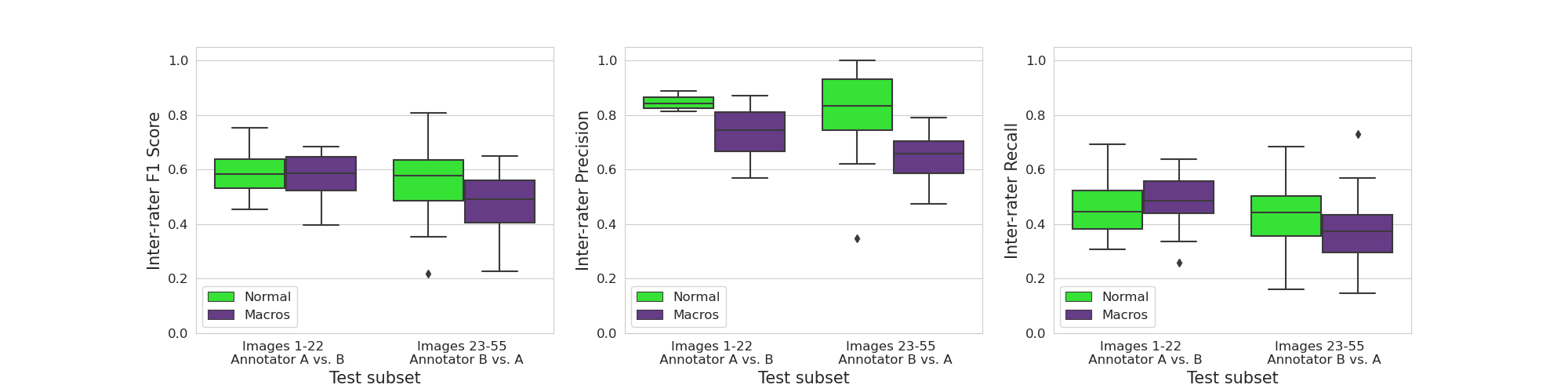}}
    \includegraphics[width=1.0\columnwidth]{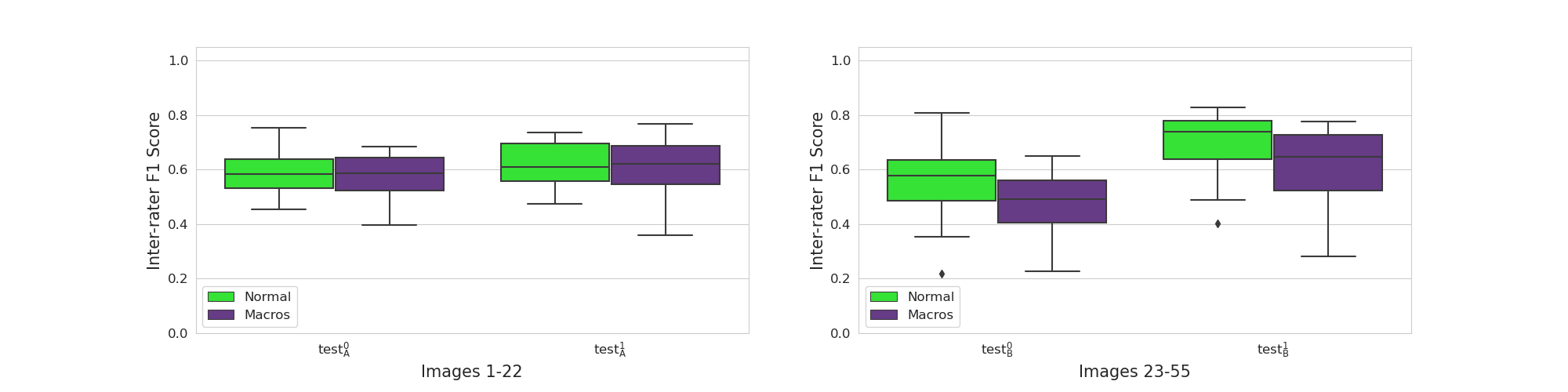}
    \caption{Inter-rater scores for the two subsets of the test set.
\textbf{Top}: F1-score (left), Precision (middle), and Recall (right) shown for Intra- and Inter-rater scores for both annotators. Scores are split for the two test subsets and according to study type. $test^0$ is always considered the \eac{GT} for computing these scores.
    \textbf{Bottom}: F1-score on images 1-22 (left), where $test^1_B$ is used as the \eac{GT} and scores are computed against $test^0_A$ and $test^1_A$, 
    and on images 23-55 (right), where $test^1_A$ is used as the \eac{GT} and scores are computed against $test^0_B$ and $test^1_B$.
    On the left, we see that when compared to a third independent label set corresponding to a different annotator, annotator A has slightly changed their style of annotation at timepoint $t^1$, slightly converging annotator B.
    On the right, we see an even bigger shift in annotation style. These results suggest that annotators exchanged best practices in annotation styles between timepoints $t^0$ and $t^1$.}
    \label{fig:inter_rater_over_time}
\end{figure}
\begin{table}[htbp!]
    \caption{Comparison of the label sets to pseudo labels.
    Precision, Recall, and F1-score computed between the pseudo labels and $test^0$, $test^1_A$, $test^1_B$ (considered as \eac{GT} in this computation).
    The pseudo labels are used as a starting point by each annotator to annotate the data.
    The F1-score shows that $test^1_B$ was curated the least, while the higher Recall values at time point $t^1$ confirm that more of the pseudo labels were removed than at $t^0$.}
    \label{tab:pseudo}
    \centering
    \begin{tabular}{@{}lccc@{}}
    \toprule
    \textbf{Metric} & \textbf{$test^0$} & \textbf{$test^1_A$} & \textbf{$test^1_B$} \\ 
    \midrule
    Precision       & 0.51              & 0.40                & 0.48                \\
    Recall          & 0.23              & 0.28                & 0.33                \\
    F1-score        & 0.32              & 0.33                & 0.39                \\
    \bottomrule
    \end{tabular}
\end{table}

\begin{comment}
\begin{table}[htbp!]
    \caption{Comparison of the label sets to pseudo labels.
    Precision, Recall, and F1-score computed between the pseudo labels and $test^0$, $test^1_A$, $test^1_B$ (considered as \eac{GT} in this computation).
    The pseudo labels are used as a starting point by each annotator to annotate the data.
    The F1-score shows that $test^1_B$ was curated the most.
    The higher Recall at time point $t^1$ confirms that more of the pseudo labels were removed than at $t^0$.
}
    \label{tab:pseudo}
    \centering
    {
        \begin{tabular}{| *{10}{c|} }
            \hline
                \diagbox{Metric}{Label set}
                & $test^0$
                & $test^1_A$
                & $test^1_B$ \\
            \hline
             Precision  & 0.51 & 0.40 & 0.48 \\
            \hline
             Recall & 0.23 & 0.28 & 0.33 \\
            \hline
           F1-score & 0.32 & 0.33 & 0.39 \\
            \hline
        \end{tabular}
        }
\end{table}
\end{comment}

\FloatBarrier
\newpage
\subsection{Model benchmark}
\label{sec:appendix-model}

\subsubsection{Training details}
For the training and validation pipeline of our benchmark the \textit{mmdetection} \citep{mmdetection} toolbox was used, a well-established open-source toolbox for object detection.
To make our dataset compatible with the toolbox, the bounding boxes were converted to the COCO format \citep{lin2015microsoft}. 
We adapted the original configuration for each model to set a number of fixed parameters for all models.
AdamW was used as the optimizer with a base learning rate of 1e-05, along with a linear learning rate scheduler.
The batch size was set to 16 and the training setup included standard image augmentations: Gaussian Blur, Random Flip, Random Shift, Random Affine, and Photometric Distortion with a probability of 0.5.
For all models, the pretrained COCO weights were used as initialization and the final layer was adapted to accommodate our single class.
All models were trained for 400 epochs and validated using the COCO metrics on $test^0$.
Training and validation were performed on an internal cluster that uses an NVIDIA A100 GPU with four cores and 40 GB VRAM.
The training time varied slightly depending on the model, but all were trained in less than 21 hours (\Cref{tab:train_test_params}). 

\begin{table}[h!]
    \caption{Training and testing of benchmark models. Train time is the duration of training the model once on the entire train set for 400 epochs. The best epoch is the epoch with the highest mAP on the test set $test^0$. GPU utilization indicates the approximate range of GPU utilization during training. Inference time is the average time per image inference using a single core.}
    \label{tab:train_test_params}
    \centering
    \resizebox{0.7\columnwidth}{!}{
        \begin{tabular}{@{}lcccc@{}}
        \toprule
        \textbf{Model} & Faster R-CNN & SSD & YOLOv3 & RTMDet \\
        \midrule
        Train time (hours) & 15 & 16 & 10 & 20 \\
        Best epoch & 68 & 86 & 27 & 323 \\
        Model size (MB) & 172 & 99 & 249 & 454 \\
        GPU utilization (\%) & 60-70 & 30-90 & 55-65 & 40-80 \\
        Inference time (seconds) & 18 & 59 & 13 & 114 \\
        \bottomrule
        \end{tabular}
    }
\end{table}

\begin{comment}
\begin{table}[h!]
    \caption{Training and testing of the benchmark models.
    Train time is the time it to train the model once on the entire train set for the full 400 epochs.
    Best epoch refers to the training epoch which yielded the highest \eac{mAP} on the test set $test^0$.
    GPU utilization is a rough range of the utilization of the four nodes of the GPU during training.
    Inference time is the average inference time for an image of the test set (using a single core).}
    \label{tab:train_test_params}
    \centering
    \resizebox{\columnwidth}{!}{
        \begin{tabular}{| *{10}{c|} }
            \hline
                \diagbox{Parameter}{Model}
                & \textit{Faster R-CNN}
                & \eac{SSD}
                & \eac{YOLOv3}
                & \eac{RTMDet} \\
            \hline
             Train time (hours)   & 15 & 16 & 10 & 20 \\
            \hline
            Best epoch    & 68 & 86 & 27 & 323 \\
            \hline
            Model size (MB) & 172 & 99 & 249 & 454 \\
            \hline
            GPU utilisation (\%) & 60-70 & 30-90 & 55-65 & 40-80 \\
            \hline
            Inference time (seconds) & 18 & 59 & 13 & 114 \\
            \hline
        \end{tabular}}
\end{table}
\end{comment}

%More specifically: \textit{Faster R-CNN} used 40-60\% of the GPUs for around 15 hours, \eac{SSD} used 20-60\% of the compute for 16 hours, \textit{YOLOv3} 40-60\% for roughly 10 hours and \eac{RTMDet} trained for around 20 hours using 20-60\% of the GPU power.

\subsubsection{Additional results}
\begin{figure}[h!!]
    \centering
    \includegraphics[width=0.9\textwidth]{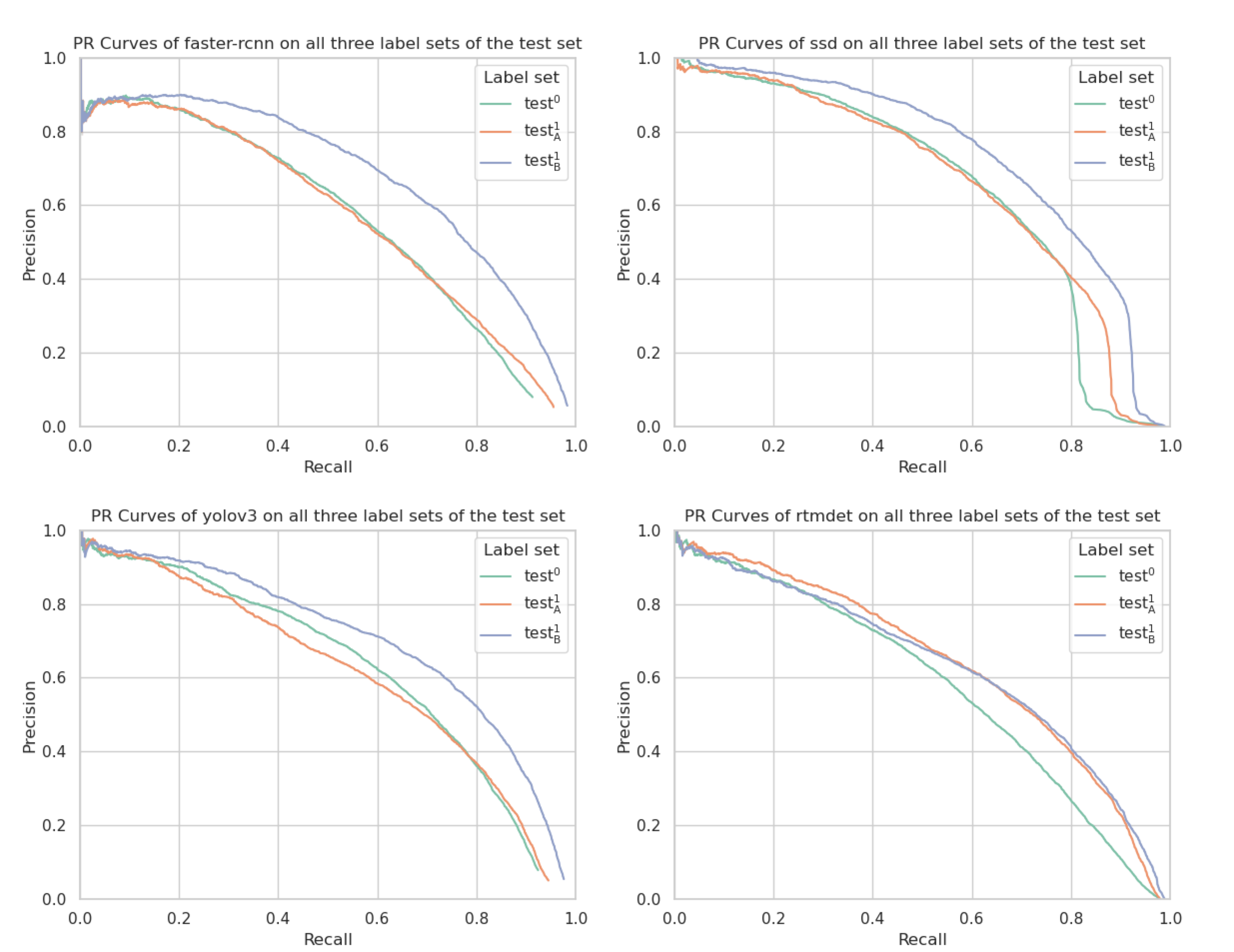}
    \caption{Model benchmark. \eac{P-R} curves across all models of the benchmark for all three label sets. It is interesting to observe that though the model checkpoints were selected based on $test^0$, they are consistently more in agreement with labels of $test^1_B$.}
    \label{fig:pr_curves_all_models}
\end{figure}
\begin{figure}[htbp!]
    \centering
    \includegraphics[width=0.9\textwidth]{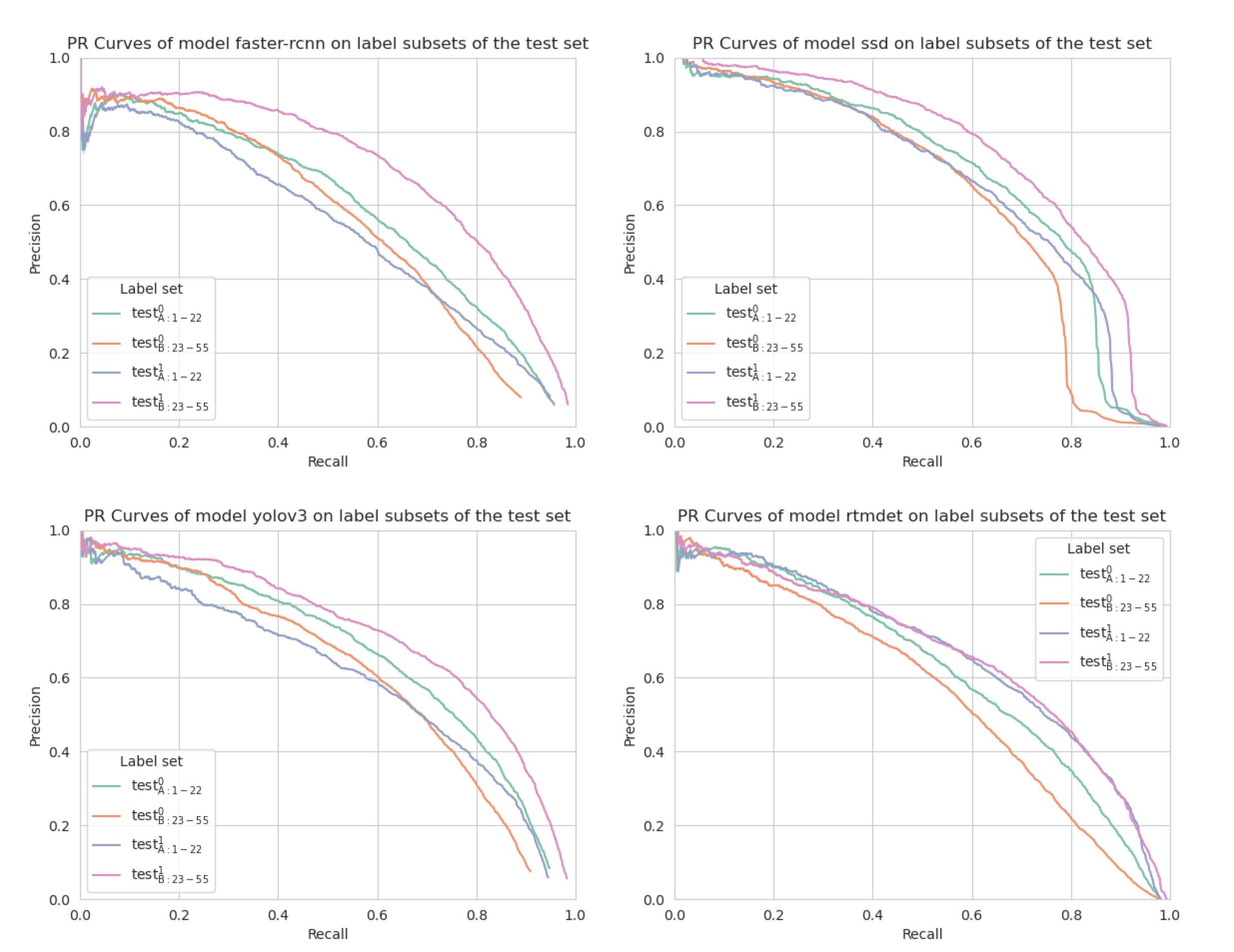}
    \caption{Model benchmark. \eac{P-R} curves across all models of the benchmark for subsets of the label sets: $test^0_{A:1-22}$ and $test^0_{B:23-55}$, along with the corresponding subsets for $test^1_A$ and $test^1_B$, i.e. $test^1_{A:1-22}$ and $test^1_{B:23-55}$ such that a direct comparison of the same subsets of the test set can be made. We see that all models are more in agreement with Annotator B at timepoint $t^1$ compared to timepoint $t^0$ for images 23-55 of the test set.}
    \label{fig:pr_curves_all_models_all_splits}
\end{figure}
\begin{figure}[h!]
    \centering
    \includegraphics[width=1.0\textwidth]{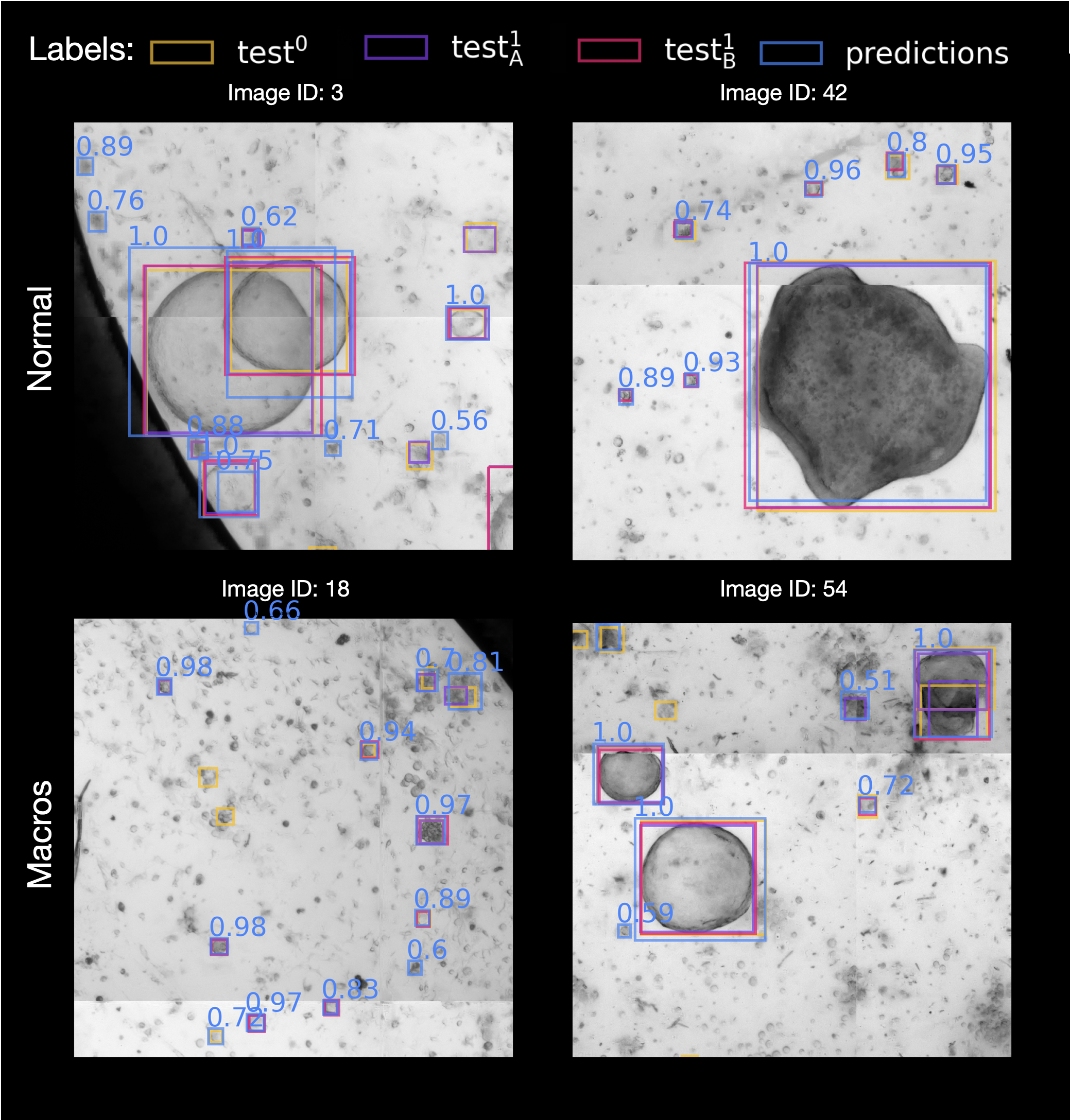}
    \caption{Example Predictions. Predicted Bounding boxes and model confidence from the \eac{SSD} model for various image crops of size 1000x1000 pixels of the 'Normal' (top) and 'Macros' (bottom) study types. The three label sets are also displayed for comparison.
    The 'Macros' images are noisier and, therefore, more challenging for the model and the annotators alike.}
    \label{fig:model_preds}
\end{figure}

\FloatBarrier
\subsection{Extensions of napari-organoid-
counter v.0.2.}
\label{sec:app-napari-plugin}

The main extensions of the latest version of the \emph{napari-organoid-counter} plugin are as follows:

\begin{itemize}
    \item Back-end: Use of the trained \eac{YOLOv3} model presented in \Cref{sec:model-training} for object detection. This model was chosen as it gives the best trade-off between performance and inference time for real-time applications.
    \item Back-end: Inference with a sliding window and adjustable parameters for multiple window sizes and window down-sampling rates.
    \item Back-end: Adjustable model confidence threshold.
    \item Front-end: Organoid ID and model confidence displayed in the viewer - the individual exported features can now be traced back to the organoids in the viewer.
    \item Front-end: Possibility to work interactively with different layers simultaneously by adjusting parameters and switching between shape layers.
\end{itemize}
\subsection{Ethics Statement}
\label{sec:ethics}

We have thoroughly reviewed this work for any potential ethical implications and believe that the societal benefits outweigh the potential issues related to this work.

The dataset we are submitting consists of lung organoids derived from murine cells. 
The data we provide is original, and any supplementary data included adheres to the Creative Commons licensing terms.
We have verified that all datasets used in our submission are current and have not been deprecated by their original creators.
Privacy-related concerns are minimal, as our dataset does not involve human subjects and, therefore, does not entail sensitive personal data, or sensitive information of humans, thereby mitigating common ethical concerns related to privacy and consent.

The dataset is intended for research within the scope of lung organoids derived from murine cells, and its applicability is limited to this specific area.
Users of the dataset are encouraged to consider this limitation and ensure their research appropriately reflects the dataset's scope and intended use when taking up our work and using it beyond the specific biological domain for which we created the dataset.
Prior to using this dataset the intended use of the model should be considered against the scope and, if the dataset is used outside the scope we refer to, it should be thoroughly tested regarding the new context to avoid biases and ensure the reliability of the resulting models.

No matter the context the dataset is used in, we would like to emphasize the importance of maintaining a human in the loop for final decision-making processes to avoid over-reliance on automated systems.

However, we nevertheless emphasize the importance of responsible use of this dataset.
Researchers utilizing this dataset should ensure that all analyses and applications are conducted within ethical guidelines and that any machine learning models or automated tools developed using this data are implemented with care.

\clearpage

%%%%%%%%%%%%%%% Datasheet for Datasets %%%%%%%%%%%%%%%%%%%%
\section{MultiOrg Datasheet}
\label{sec:datasheet}
This data sheet serves as supplementary documentation aimed at improving reproducibility. It is based on the guidelines outlined in \textbf{Datasheets for Datasets}\footnote{\url{https://arxiv.org/abs/1803.09010}}, a working paper developed for the machine learning community.

\begin{itemize}

\item \textbf{For what purpose was the dataset created?}
%To foster research development in machine learning for graphs, in particular its application to neuroscience - specifically the brain vessel graph composition.
This dataset was created with two goals in mind: a. to facilitate research in uncertainty quantification methods in machine learning and b. to enable the development of object detection models for the automated detection of lung organoids, which can accelerate the annotation process for similar data in future studies.

\item \textbf{Who created the dataset (e.g., which team, research group) and on behalf of which entity (e.g., company, institution, organization)?}
%The dataset was created thorough a collaborative effort by neuro-scientists and computer-scientists at the Technical University of Munich and the Helmholtz Zentrum München (under the supervision of Ali Ertuerk, Bjoern Menze and Stephan Günnemann).
The dataset was created as a collaborative effort by the authors of this work, i.e. biologists of the Institute of Lung Health and Immunity (LHI) of the Helmholtz Zentrum München and the Philipp University of Marburg, as well as computer scientists Helmholtz AI at the Helmholtz Zentrum München.
  
\item \textbf{Who funded the creation of the dataset?} 
%The creation of the dataset was funded only indirectly via the salaries of the scientists at the Technical University of Munich and the other corresponding affiliations of the authors. 
This work was partly funded by the German Center for Lung Research (DZL) and from the Deutsche Forschungsgemeinschaft (DFG, German Research Foundation– 512453064) as well as from the Stiftung Atemweg.
Additionally, it was developed as part of the daily work of the creators and was indirectly funded via their salaries.

\end{itemize}

\subsection{Composition}

\begin{itemize}

\item \textbf{What do the instances that comprise the dataset represent?} 
Our dataset consists of 2D images of microscopy plate wells, consisting of lung organoids derived from murine cells.
Along with the imaging data, annotations for each image are provided, in the form of bounding boxes fit around the organoids, and metadata information on the annotator and time point of annotation. 
%Our dataset represents graph representations of the whole brain vasculature. We are providing two alternative representations of the vascular graph. First, a representation where individual vessels are represented as edges in a Graph; and second, the corresponding Line graph were vessels are represented as nodes. One can interpret the graph of a single mouse brain as a single instance. Alternatively one can interpret each vessel (edge) and bifurcation (node) as a physical instance. 

\item \textbf{How many instances are there in total (of each type, if appropriate)?}
In total 411 fully annotated images are released as part of this dataset, annotated by two annotators at two different timepoints and consisting of 26 different experiments across two biological study setups.
Please see \Cref{dataset} and \Cref{tab:data-overview} for more details and stratification between study types, annotators, train and test splits.

%By the instance definition of whole brain graphs as instances we are providing 17 graphs (with the option to generate the line graph) from 3 imaging sources as instances. In the future we plan to extend the dataset as soon as other whole brain vessel segmentations are made publicly available (open source).  

%By the definition of vessels and bifurcation points we have millions of instances for each. Please see Table \ref{tab:graph_numbers_nodes_edges} for detailed numbers.

\item \textbf{Does the dataset contain all possible instances or is it a sample of instances from a larger set?} 
%We are providing all available instances. 
The dataset is a subset of a larger set of biological experiments. During the first curation of the data, the 26 experiments were selected to have the best representation of our data, with the least noise stemming from the image acquisition and with an acceptable number of organoids in the well, neither too many, which would make it hard to distinguish them from one another, nor too few, with little information present in the image.

\item \textbf{What data does each instance consist of?} 
%In either case, please provide a description. By definition 1); each instance represents a whole mouse brains' vascular graph saved in the widely used CSV format. By definition 2) each node represents a bifurcation point and each edge a vessel. 
Each instance is an image of size between 5719 and 6240 pixels in the \textit{x} and 5551 and 6940 pixels in the \textit{y} axis respectively. 
Each pixel in the image is equivalent to 1.29 \(\mu m\) in each axis.

\item \textbf{Is there a label or target associated with each instance?}
%Yes, in case of the edge and node instances, the information from the extracted graphs (features) can be used as the instance labels. E.g. in our node classification benchmark we use the vessel radius binned in three classes as an instance label. 
Yes, specifically for the training set one target per image is available, while for the images of the test set, we release three sets of labels. The first, which in this work is named $test^0$, is directly available, while the other two can be indirectly accessed by participating in our \href{https://www.kaggle.com/competitions/multi-org-challenge}{Kaggle competition} and submitting results to our leaderboard.
  
\item \textbf{Is any information missing from individual instances?}
%No, all of the information has been provided.
No, to the best of our knowledge, all available information has been provided.

\item \textbf{Are relationships between individual instances made explicit?} 
%In our dataset the instances (brain graphs) are independent.
Yes, \Cref{tab:data-overview} and the structure of the available data make relationships explicit.  Relationships become apparent with the data structure provided in the \href{https://www.kaggle.com/datasets/christinabukas/mutliorg?select=croissant_mutliorg_metadata.json}{metadata file}. For example, all image wells belonging to the experiment "Plate\_1" can be found in a folder named "Plate\_1".

\item \textbf{Are there recommended data splits (e.g., training, development/validation, testing)?} 
Yes, we provided the data already split into train and test sets.
As discussed in the main manuscript, for validation we use the test data with one of the three label sets.
%For the benchmark we split one whole brain into a train, validation and test set of 80/10/10.
  
\item \textbf{Are there any errors, sources of noise, or redundancies in the dataset?}
%Our graph extraction is based on experimental imaging and segmentation techniques. Therefore, errors and uncertainty are inherent. We discuss these in detail in our Limitations section in the conclusion. 
Noise is always present in microscopy data, and is one of the reasons for which machine learning tasks in the biomedical domain are much harder compared to natural images.
This noise can derive from the microscope itself, the imaging parameters, or the biological specimen. 
Nevertheless, we tried to eliminate these as much as possible when curating the dataset, by, as mentioned above, selecting our experiments out of a larger pool and 
Another source of noise in our case is the stitching of the imaging tiles to form the 2D image well, which we discuss in \Cref{dataset}.
No redundancies are present in the dataset.

\item \textbf{Is the dataset self-contained, or does it link to or otherwise rely on external resources (e.g., websites, tweets, other datasets)?} 
The provided dataset is self-contained. 

\item \textbf{Does the dataset contain data that might be considered confidential (e.g., data that is protected by legal privilege or by doctor-patient confidentiality, data that includes the content of individuals' non-public communications)?} 
No.

\item \textbf{Does the dataset contain data that, if viewed directly, might be offensive, insulting, threatening, or might otherwise cause anxiety?} 
No. 

\item \textbf{Does the dataset relate to people?} 
No. 

\end{itemize}

\subsection{Collection process}

\begin{itemize}

\item \textbf{How was the data associated with each instance acquired?} %The data was generated from a set of different publicly available datasets of whole murine brain images and segmentations. The specifics of the generation of each of these public segmentations are specified in the referenced literature and their licenses, see \ref{Individual_Licenses_Data}. 
The data was acquired by cell culture and imaging in a Life Cell Imaging Microscope.
    
\item \textbf{What mechanisms or procedures were used to collect the data?} %We use the \textit{Voreen} framework \citep{drees2021scalable,meyer2009voreen} to generate graphs from segmentations. \textit{Voreen} is a software which runs on a CPU. 
The biological experiments consisted of isolating and culturing murine cells, which received various treatments and formed variable organoids in the process.
Subsequently, images were taken to document and analyze the effects of different treatments. 

\item \textbf{If the dataset is a sample from a larger set, what was the sampling strategy?} %The dataset is complete. 
The dataset is a subset of a larger set of biological experiments.
During the first curation of the data, the 26 experiments were selected to have the best representation of our data, with the least noise stemming from the image acquisition and with an acceptable number of organoids in the well, neither too many, which would make it hard to distinguish them from one another, nor too few, with little information present in the image.

\item \textbf{Who was involved in the data collection process (e.g., students, crowdworkers, contractors) and how were they compensated (e.g., how much were crowdworkers paid)?}
%Only researchers (co-authors) of the Technical University of Munich and the Helmholtz Zentrum München were involved in the data collection process.
The data was collected by university students with guest contracts and employees of the Helmholtz Zentrum Munich.

\item \textbf{Over what timeframe was the data collected?} %Does this timeframe match the creation timeframe of the data associated with the instances (e.g., recent crawl of old news articles)?  The generation of the dataset, including dedicated research to gather the base segmentations and to optimize the graph extraction procedure took roughly one year. 
The data was collected at specific timepoints over a period of two weeks. All experiments took place over a timeframe of two years.

\item \textbf{Were any ethical review processes conducted (e.g., by an institutional review board)?} No
%Our work is purely based on public and open sourced data. However, ethical review processes were carried out for each of these open sourced base segmentation sets:

%The three graphs from Ji et al. \citep{ji2021brain} are based on animal experiments,  they followed the Guide for the Care and Use of Laboratory Animals and have been approved by the Institutional Animal Care and Use Committee, for details see \url{https://doi.org/10.1016/j.neuron.2021.02.006}.

%The animal experiments for the nine datasets from the VesSAP paper \citep{todorov2020machine} were carried out under approval of the ethical review board of the government of Upper Bavaria (Regierung von Oberbayern, Munich, Germany), and in accordance with European directive 2010/63/EU for animal research, for details see \url{https://doi.org/10.1038/s41592-020-0792-1}. 

\item \textbf{Does the dataset relate to people?} No, it is a murine dataset.

\end{itemize}

\subsection{Preprocessing/cleaning/labeling}

\begin{itemize}

\item \textbf{Was any preprocessing/cleaning/labeling of the data done ?} %Yes, this actually constitutes a core contribution of our work, therefore please refer to Section \ref{Graph_Ext} in the main paper and to Supplementary section \ref{Suppl_Graph_Doc}.
The data was indeed preprocessed, cleaned, and labeled. In the Appendix, we describe these processes under \Cref{sec:annotation_procedure} and \Cref{sec:image_acquisition}.

\item \textbf{Was the 'raw' data saved in addition to the preprocessed/cleaned/labeled data (e.g., to support unanticipated future uses)?} %The raw data are the base segmentations. They are publicly available, the links are provided in Supplementary section \ref{Individual_Licenses_Data}.
Yes, the raw data was saved and stored.

\item \textbf{Is the software used to preprocess/clean/label the instances available?}
%The \textit{Voreen} software used for the graph extraction is publicly available, see our github repo.
The software used for preprocessing the data, \emph{Zen 2 Blue} software by \emph{Carl Zeiss Microscopy GmbH}, is proprietary.
The software used for annotating the data of the \emph{napari-organoid-counter} tool \citep{christina_bukas_2022_7065206}, a plugin developed for \emph{Napari}, is open-source and freely available.

\end{itemize}

\subsection{Uses} 
\begin{itemize}
\item \textbf{Has the dataset been used for any tasks already?} %In its current size and level of labeling detailization, the dataset was not used before (besides for the presented link prediction and node classification in this work). 
The dataset was used as part of this work to create the benchmark presented in \Cref{sec:model}.
The trained models are also made publicly available through this work and can be accessed on \href{https://zenodo.org/records/11258022?}{zenodo}.
Moreover, the latest version of the \emph{napari-organoid-counter} tool described in \Cref{sec:napari-plugin} also uses one of the models from this benchmark, trained with the current dataset.

\item \textbf{Is there a repository that links to any or all papers or systems that use the dataset?} If so, please provide a link or other access point.

No, please refer to \Cref{sec:availability} instead.

\item \textbf{What (other) tasks could the dataset be used for?} 
We discussed previously how the intended usage of our dataset is two-fold, a. to facilitate research in uncertainty quantification methods in machine learning and b. to enable the development of object detection models for the automated detection of lung organoids, which can accelerate the annotation process for similar data in future studies.
Aside from these tasks, one could use this dataset to benchmark new model architectures for object detection, or to develop unsupervised learning methods for organoid classification (using the bounding boxes to extract single organoid images), since we mention that our dataset consists of two different types of organoids. 

%In the main paper we discussed two standard tasks in machine learning on graphs; we think that our dataset can serve as a starting point for many interesting research directions in machine learning research and neurovascular research. 

\item \textbf{Is there anything about the composition of the dataset or the way it was collected and preprocessed/cleaned/labeled that might impact future uses?} 
No. 

\item \textbf{Are there tasks for which the dataset should not be used?} Not to the best of our knowledge.

\end{itemize}

\subsection{Distribution}

\begin{itemize}

\item \textbf{Will the dataset be distributed to third parties outside of the entity (e.g., company, institution, organization) on behalf of which the dataset was created?} %Our Dataset is open sourced under a CC Attribution-NonCommercial 4.0 International (CC BY-NC 4.0) License. Therefore all third parties can openly access it. 
Yes, the dataset is hereby made publicly available under the CC BY-NC-SA 4.0 License, and can therefore be used by third parties.

\item \textbf{How will the dataset will be distributed (e.g., tarball  on the website, API, GitHub)?} 
The dataset is hereby made available via the Kaggle platform and can be accessed through the link: https://www.kaggle.com/datasets/christinabukas/mutliorg/ and DOI: 10.34740/kaggle/ds/5097172
%Yes, our DOI is \url{10.5281/zenodo.5301621}

\item \textbf{When will the dataset be distributed?} %The dataset is available from the moment of submission. 
The dataset is made available along with the submission of the current manuscript.

\item \textbf{Will the dataset be distributed under a copyright or other intellectual property (IP) license, and/or under applicable terms of use (ToU)?} %Our Dataset is open sourced under a CC Attribution-NonCommercial 4.0 International (CC BY-NC 4.0) License.
Our dataset is open source and made available under the CC BY-NC-SA 4.0 License.

\item \textbf{Have any third parties imposed IP-based or other restrictions on the data associated with the instances?} No. 

\item \textbf{Do any export controls or other regulatory restrictions apply to the dataset or to individual instances?} No.

\end{itemize}

\subsection{Maintenance}

\begin{itemize}

\item \textbf{Who is supporting/hosting/maintaining the dataset?}
The dataset is maintained by the authors of this work.
The dataset is currently hosted on the Kaggle\footnote{\url{https://www.kaggle.com}} platform. 

%The dataset is initially supported and maintained by the lead authors of this paper. The data is initially hosted on a university server and links are provided in the github repository \url{https://github.com/jocpae/VesselGraph}. In the long term we aim to incorporate our dataset into the open graph benchmark (OGB) initiative\footnote{\url{https://ogb.stanford.edu/}}. 

\item \textbf{How can the owner/curator/manager of the dataset be contacted (e.g., email address)?} %Of course via e-mail:  \url{johannes.paetzold@tum.de} and via the github repository, see question above. 
Kaggle offers a discussion tab under the dataset repository.
This can be used for any data-related discussions, while there is also a discussion tab available on the website of our competition.
Naturally, the corresponding authors of this work may also be contacted directly with any questions via email.

\item \textbf{Is there an erratum?} %At this stage no, but we are happy to track them in a dedicated file in our github repository. 
There is currently no erratum for this dataset.

\item \textbf{Will the dataset be updated (e.g., to correct labeling errors, add new instances, delete instances)?} %Yes, we release the dataset on open platforms on which we plan to continuously update our dataset. Particularly to add novel whole brain vessel graphs to the dataset.
There are currently no immediate plans for updating the dataset.
We are eager to first see how it will be accepted by the community and which needs will arise for future versions/extensions. We, of course, plan to maintain the dataset, e.g. if errors are found in the data, we will update the dataset and release a newer version.
    
\item \textbf{If the dataset relates to people, are there applicable limits on the retention of the data associated with the instances?} The dataset does not relate to people. 
    
\item \textbf{Will older versions of the dataset continue to be supported/hosted/maintained?} 
If and when newer versions of the dataset are released we expect these to be an improvement upon the original version, and will therefore concentrate our efforts on maintaining the latest version of the dataset.
%When novel versions of the dataset will be released we will continue to host and maintain the old versions of the dataset.
    
\item \textbf{If others want to extend/augment/build on/contribute to the dataset, is there a mechanism for them to do so?} 
Researchers are more than welcome to extend our dataset.
In the discussion section of our manuscript, \Cref{sec:discussion}, we mention how future versions of our dataset could include even more label sets both for the train and test sets.
Such versions would surely increase its value and the development of uncertainty quantification techniques.

%We encourage other researches to exactly that. Depending on their contribution they can contribute to our github repository (in case of implementations) or reach out to us via e-mail in case they want to contribute graphs to the dataset. Our dataset and code are open sourced, see above. 

\end{itemize}

\newpage

\section{Availability: data, benchmark, and software tool}
\label{sec:availability}
Below we list all assets made publicly available with the release of this work:
\begin{itemize}
    \item Trained model weights from our benchmark can be found on \href{https://zenodo.org/records/11258022?}{zenodo} with a DOI: 10.5281/zenodo.11258022
    \item The MultiOrg dataset can be found on \href{https://www.kaggle.com/datasets/christinabukas/mutliorg/}{Kaggle} with a DOI: 10.34740/kaggle/ds/5097172
    \item The notebooks for reproducing our benchmark can be found under the same repository on the \href{https://www.kaggle.com/datasets/christinabukas/mutliorg/code}{Kaggle MultiOrg dataset page}
    \item The Croissant metadata record documenting the dataset can also be found on the Kaggle MultiOrg dataset page \href{https://www.kaggle.com/datasets/christinabukas/mutliorg?select=croissant_mutliorg_metadata.json}{here} 
    \item The napari plugin can be found on the \href{https://www.napari-hub.org/plugins/napari-organoid-counter}{napari-hub}
\end{itemize}

\newpage

\section{Author Statement}

As the authors of this dataset, we hereby declare that we bear full responsibility for any and all consequences arising from the use, distribution, and publication of the dataset.
This includes but is not limited to, any violations of privacy, intellectual property rights, or any other legal rights.

We confirm that all data included in this dataset has been collected, processed, and shared in compliance with applicable laws and regulations.
We have obtained all necessary permissions and consents from individuals or entities involved, and we affirm that the data does not infringe upon the rights of any third parties.

Furthermore, we confirm that the dataset is being released under the following license:  CC BY-NC-SA 4.0. 
his license allows others to use, share, and adapt the dataset, provided that appropriate credit is given, a link to the license is provided, and any changes are indicated.

By submitting this dataset to the NeurIPS 2024 dataset track, we agree to adhere to the terms and conditions set forth by the NeurIPS conference organizers and acknowledge that we are solely responsible for any issues related to the dataset's legal and ethical use.

\newpage

\section{Hosting, licensing, and maintenance plan.}
\label{sec:hosting}

As described in the Maintenance section of \Cref{sec:datasheet}, the dataset is open source and available to researchers via the Kaggle\footnote{\url{https://www.kaggle.com}} platform, under the CC BY-NC-SA 4.0 license. Additionally, on kaggle we offer notebooks to reproduce the model benchmark performed in this work, and links to our zenodo repository which stores our pretrained models. All notebooks are under the Apache 2.0 license and model weights are available under the Creative Commons Attribution 4.0 International license.
All the above, will be maintained actively by the authors of this work.
If errors are found by users in the code or data, this can be communicated via the discussion tab available on the dataset webpage, and a newer version will be uploaded.

\end{document}